\documentclass[11pt]{article}

\usepackage[preprint]{acl}

\usepackage{times}
\usepackage{latexsym}

\usepackage[T1]{fontenc}

\usepackage[utf8]{inputenc}

\usepackage{microtype}

\usepackage{inconsolata}

\usepackage{graphicx}

\usepackage[utf8]{inputenc} 
\usepackage[T1]{fontenc}    
\usepackage{hyperref}       
\usepackage{url}            
\usepackage{booktabs}       
\usepackage{amsfonts}       
\usepackage{nicefrac}       
\usepackage{microtype}      
\usepackage{xcolor}         
\definecolor{lightred}{RGB}{255,200,200}
\definecolor{lightblue}{RGB}{191,224,250}
\usepackage{graphicx}
\usepackage{mathrsfs}
\usepackage{amsmath}
\usepackage{wrapfig}
\usepackage{tabularx}
\usepackage{amsthm}
\usepackage{amssymb}
\usepackage{booktabs}
\usepackage{algorithm}
\usepackage{algorithmic}
\usepackage{booktabs}
\usepackage{pifont}
\usepackage{colortbl}
\usepackage[most]{tcolorbox} 
\usepackage{enumitem}
\usepackage{longtable}
\usepackage{multirow}

\title{Cognitive Enhancement: Rethinking the Necessity of Role-Playing for Large Language Models}

\author{
  Xingjie Zhuang$^{1,\dagger}$, Jialong Tang$^{2,\dagger}$, Chulun Zhou$^{3,}$\thanks{Corresponding authors.}, Buchao Zhan$^{1}$, Zhirui Li$^{1}$,\\  \textbf{Junhui Li}$^{4}$, \textbf{Yazheng Yang}$^{1}$, \textbf{Jinsong Su}$^{1,}$\footnotemark[1] \\
  $^{1}$School of Informatics, Xiamen University,\\ $^{2}$Tongyi Lab, $^{3}$The Chinese University of Hong Kong, $^{4}$Soochow University,\\
\texttt{zhuangxj@stu.xmu.edu.cn; jssu@xmu.edu.cn} \\
}

\begin{document}
\maketitle

\insert\footins{\footnotesize$^\dagger$Equal contribution.}
\begin{abstract}
Role-playing prompting has become a popular yet simple technique for improving LLM reasoning and output quality. However, whether it consistently boosts performance across diverse domains remains unclear, as systematic validation is lacking. To fill this gap, we run multi-model, cross-domain, and multilingual experiments on MMLU and MMLU-Redux. We find that gains from role-play prompting depend heavily on model capacity, knowledge domain, and prompt language. Drawing on metacognition theory, we propose the persona-related cognitive alignment hypothesis: role-play works only when the LLM correctly grasps the designated persona and its associated knowledge domain. We test this hypothesis through persona information richness ablation, layer-wise entropy divergence analysis, and latent thought-space deflection observation. To reduce persona cognitive bias and stabilize role-play performance, we propose \textbf{M}ixed-\textbf{L}anguage \textbf{C}oncatenate \textbf{P}rediction \textbf{(MLCP}), a simple, training-free, and efficient multilingual prompt concatenation strategy. It aggregates semantically equivalent role prompts to enrich complementary representational cues. Extensive experiments show that MLCP consistently outperforms vanilla role-play prompting across all tested LLMs. Our code and implementation details are released at \textcolor{blue}{\url{https://github.com/XMUDeepLIT/RethinkingRolePlay}}

\end{abstract}

\section{Introduction}

Recently, Large Language Models (LLM) have demonstrated remarkable capabilities in complex reasoning, knowledge-based question answering (QA), and domain-specific tasks \cite{ref2,ref3,ref1}. As model capabilities evolve, researchers have increasingly found that carefully designed prompts can significantly modulate model behavior. Among these paradigms, role-playing prompting, which assigns specific identities, professions, or expert personas to models, has emerged as a widely adopted strategy. It has proven effective in enhancing performance across a myriad of scenarios, including reasoning, QA, decision support, and autonomous agents. Consequently, implementing persona-based prompts such as "Please answer the following question as an expert in [Domain]" has become a standard practice in current LLM applications.

\begin{figure}[t!]   
	\centering
 \includegraphics[width=1
\linewidth]{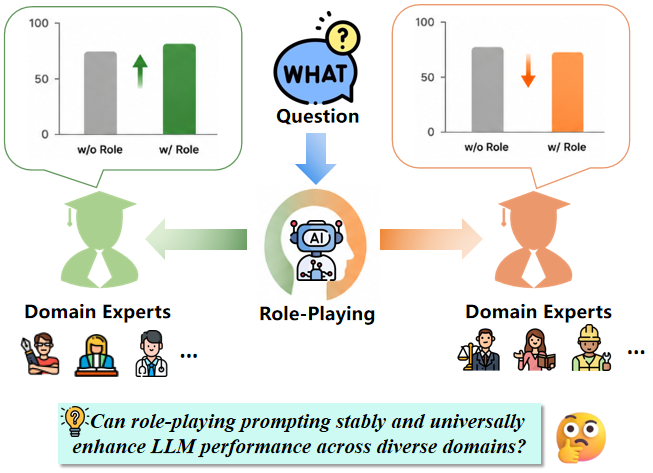}
	\caption{Performance comparison of domain experts with and without role-playing prompts.}
	\label{Figure1}
\end{figure}

However, existing research work predominantly focus on the success cases of role-playing in isolated tasks or specific scenarios, implicitly assuming that role-playing prompts can consistently and unconditionally boost model performance \cite{ref4,ref30,ref7,ref5,ref6}. As visually illustrated in Figure \ref{Figure1}, this common assumption does not always hold true. A critical question that remains under-explored is: \textit{\textbf{Can role-playing prompting stably and universally enhance LLM performance across diverse domains?}} In other words, does role-playing prompting function as a universally beneficial capacity-enhancement mechanism, or is it merely a highly conditional task-adaptation strategy?

To systematically investigate this question, we first conducted comprehensive preliminary study to evaluate the impact of role-playing on knowledge-based QA capabilities across multiple mainstream LLMs. Utilizing two extensive multidisciplinary benchmark datasets, MMLU \cite{ref8} and MMLU-Redux \cite{ref9}, we analyzed the experimental outcomes across three dimensions: overall performance, subject-level comparison, and language form. The empirical results reveal three prominent and intriguing phenomena:

\textbf{First}, in terms of overall performance, role-playing does not yield consistent gains. Models differ substantially in how they respond to role prompts: stronger models tend to benefit more, while weaker ones often gain little or even degrade. This trend is confirmed within the same model series, where improvements rise with model scale. These results suggest that role-playing is not a universally effective strategy and that its effectiveness is closely tied to model capability.

\textbf{Second}, subject-level analysis shows that the benefits of role-playing vary greatly across knowledge domains. Even models from the same family but with different parameter scales can exhibit opposite performance changes on the same subject, and such discrepancies are even more pronounced across different model families. Thus, role-playing does not produce steady gains in any individual subject; its effectiveness is highly conditional, jointly shaped by question type, knowledge domain, and model characteristics.

\textbf{Finally}, we examine how prompt language affects role-playing performance. Simply changing the language of a role prompt can lead to significantly different outcomes. Notably, even for large-scale models pretrained on massive English and Chinese data, the best performance does not necessarily occur with role prompts in these high-resource languages. For example, Qwen3 performs worse under Chinese role prompts than under certain alternative language variants. These findings suggest that role-playing effectiveness depends not only on the assigned role itself, but also on the prompt language and the knowledge cognition it activates within the model.

Together, these findings show that the widely adopted role-playing strategy is far more intricate than previously assumed. It is neither a universally stable enhancement mechanism nor does it have a standardized optimal template, and its underlying mechanism remains poorly understood. Relying solely on empirical persona design can easily fail to ensure reliable performance gains.

Driven by these observations, we further investigate the internal mechanisms behind the performance gains and degradations of role-playing prompting. We first analyze several experimental phenomena, including variations in persona information richness, internal entropy fluctuations, and latent thought-space deflections. Inspired by meta-cognition \cite{ref28}, we propose the persona-related cognitive alignment hypothesis to explain the working mechanics of role-playing and persona cognitive biases, and comprehensively validate it. Building on these insights, we introduce Mixed-Language Concatenate Prediction (MLCP), a simple, training-free, and efficient multilingual prompt concatenation strategy that aggregates semantically equivalent role prompts to exploit complementary representational cues and mitigate role-play prompt failures. Experiments confirm that MLCP delivers robust and stable performance improvements across various models and task scenarios.

Our contributions are threefold:
\begin{enumerate}
\item We perform the first systematic evaluation of role-playing in knowledge QA covering extensive diverse subjects across different types of LLMs, benchmarks and prompt languages, and build a comprehensive analytical pipeline to evaluate the effectiveness of role-playing.

\item Multi-dimensional experiments verify that role-playing yields no universal improvements in knowledge QA. Its performance heavily depends on model capability, subject domain and prompt language, demonstrating strong conditionality. We further propose and fully validate the persona-related cognitive alignment hypothesis.

\item We explain how multilingual settings impact role-playing via cognitive space deflection and displacement, and introduce the training-free MLCP strategy to reduce persona cognitive bias. Extensive experiments confirm that MLCP yields steady performance gains as an efficient and lightweight prompt optimization method.
\end{enumerate}
\section{Preliminary Study}

\subsection{Study Setup}
\label{section:Study Setup}
We evaluate the model under two distinct prompting paradigms: Base Prompt ($\mathcal{P}$$_{\text{base}}$) and Role-Playing Prompt ($\mathcal{P}$$_{\text{rp}}$). While $\mathcal{P}$$_{\text{base}}$ presents the benchmark instances directly to the model, $\mathcal{P}$$_{\text{rp}}$ encapsulates the question within a specific role-playing persona instruction. Considering that users in real-world utilization typically instantiate role-playing prompts in their respective native languages even when querying the same core problem, these persona instructions are implemented across a multilingual set $\mathcal{L} = \{\text{en}, \text{zh}, \text{es}, \text{fr}, \text{ja}, \text{de}, \text{km}, \text{ms}\}$, ensuring that the underlying semantic meaning remains invariant across all languages $l \in \mathcal{L}$.

Formally, we define a formatting function $f(\cdot)$ that unifies the prompt template, benchmark instance, and language-specific evaluation configuration to form the complete model input. For any given question instance, the model generates its optimal predicted answer $\hat{y}$ by maximizing the conditional probability of candidate options given the formatted prompt input. The prediction process can be formulated as:
\begin{equation}
\hat{y} = \arg\max_{c \in C} P\big(c \mid f(Q, C, \mathcal{P}, l); \mathcal{M}\big)
\end{equation}

Accordingly, the accuracy $\mathcal{A}$ for a model $\mathcal{M}$ on a specific subject $s$ under a given configuration is defined as:
\begin{equation}
\mathcal{A}_{\text{s}}(\mathcal{M}, \mathcal{D}_s, \mathcal{P}, l) = \frac{1}{|\mathcal{D}_s|} \sum_{(Q, C, y^\star) \in \mathcal{D}_s} \mathbb{I}\left(\hat{y} = y^\star\right)
\end{equation}
where $\mathbb{I}(\cdot)$ represents the indicator function that outputs 1 if the condition holds true and 0 otherwise. 

To capture performance at a macro level, the overall average accuracy across the entire benchmark \(\mathcal{D}\) is computed as the mean performance over all constituent subjects:
\begin{equation}
\mathcal{A}_{\text{overall}}(\mathcal{M}, \mathcal{D}, \mathcal{P}, l)
= \frac{1}{|\mathcal{S}|} \sum_{s\in\mathcal{S}} \mathcal{A}_{\text{s}}(\mathcal{M}, \mathcal{D}_s, \mathcal{P}, l)
\end{equation}

\subsection{Observations}
Drawing from the empirical findings in Table \ref{table:1} and Table \ref{table:2}, we summarize several noteworthy phenomena as follows:

\textit{\textbf{Phenomenon 1: Role-playing benefits correlate with model capability.}}
Analysis of average accuracies and the count of improved versus degraded subjects reveals that stronger models benefit more substantially from role-playing. Conversely, less capable models yield diminishing or even marginal returns. This pattern is clearly verified within the same model series, where performance improvements gradually rise as the model scale increases.

\textit{\textbf{Phenomenon 2: Role-playing gains lack universal generalizability and are highly domain-dependent.}}
Subject-level analysis shows no guaranteed boost in any specific domain. Even within the same model family, varying scale or architecture leads to divergent outcomes on identical subjects, indicating that role-playing efficacy is highly conditional—jointly determined by question type and domain.

\textit{\textbf{Phenomenon 3: Linguistic phrasing in role-playing prompts significantly impacts performance gains.}} Evaluating prompts across different languages suggests that the language choice alters final gains. Remarkably, even models pre-trained on massive English and Chinese corpora do not consistently optimalize in these dominant languages (e.g., Qwen3 underperforms in Chinese compared to other variations).
\begin{table*}
\centering
\resizebox{1\linewidth}{!}{
\begin{tabular}{llcccccccccc}
\toprule
\multirow{2}{*}{Models} & \multirow{2}{*}{Benchmark} & \multirow{2}{*}{Base} & \multicolumn{9}{c}{Role-playing} \\
\cmidrule{4-12}
& & & en & zh & es & fr & ja & de & km & ms & avg (non-en) \\
\midrule
\multirow{2}{*}{Qwen3-4B} 
& MMLU-Redux & 0.756 & 0.754 & \cellcolor{lightred}{0.730} & 0.746 & 0.746 & \cellcolor{lightblue}{0.757} & 0.746 & 0.746 & \cellcolor{lightblue}{0.757} & 0.747 \\
& MMLU & 0.756 & 0.754 & \cellcolor{lightred}{0.724} & 0.762 & \cellcolor{lightblue}{0.763} & 0.757 & \cellcolor{lightblue}{0.763} & 0.757 & 0.757 & 0.755 \\
\midrule
\multirow{2}{*}{Qwen3-8B}
& MMLU-Redux & 0.764 & 0.767 & \cellcolor{lightred}{0.749} & 0.769 & 0.769 & 0.772 & 0.769 & 0.769 & \cellcolor{lightblue}{0.774} & 0.767 \\
& MMLU & 0.782 & 0.782 & \cellcolor{lightred}{0.756} & \cellcolor{lightblue}{0.796} & 0.793 & 0.792 & 0.792 & 0.788 & 0.790 & 0.787 \\
\midrule
\multirow{2}{*}{Qwen3-14B}
& MMLU-Redux & 0.798 & 0.810 & \cellcolor{lightred}{0.787} & 0.810 & 0.810 & \cellcolor{lightblue}{0.820} & 0.810 & 0.810 & 0.806 & 0.808 \\
& MMLU & 0.822 & 0.824 & \cellcolor{lightred}{0.803} & \cellcolor{lightblue}{0.829} & 0.828 & 0.823 & \cellcolor{lightblue}{0.829} & 0.819 & \cellcolor{lightblue}{0.829} & 0.823 \\
\midrule
\multirow{2}{*}{Qwen3-32B}
& MMLU-Redux & 0.834 & 0.852 & \cellcolor{lightred}{0.834} & 0.850 & 0.850 & \cellcolor{lightblue}{0.858} & 0.850 & 0.848 & 0.854 & 0.849 \\
& MMLU & 0.841 & 0.856 & \cellcolor{lightred}{0.838} & \cellcolor{lightblue}{0.858} & 0.856 & 0.854 & \cellcolor{lightblue}{0.858} & 0.850 & \cellcolor{lightblue}{0.858} & 0.853 \\
\midrule
\multirow{2}{*}{Llama3-8B}
& MMLU-Redux & 0.539 & 0.524 & 0.532 & 0.485 & \cellcolor{lightblue}{0.540} & 0.515 & 0.530 & \cellcolor{lightred}{0.249} & 0.538 & 0.484 \\
& MMLU & 0.611 & \cellcolor{lightblue}{0.618} & 0.585 & 0.589 & 0.594 & 0.548 & 0.564 & \cellcolor{lightred}{0.333} & 0.583 & 0.542 \\
\midrule
\multirow{2}{*}{Mistral-7B}
& MMLU-Redux & 0.458 & 0.467 & 0.493 & 0.467 & 0.493 & 0.491 & \cellcolor{lightblue}{0.494} & \cellcolor{lightred}{0.412} & 0.475 & 0.475 \\
& MMLU & 0.588 & \cellcolor{lightblue}{0.589} & 0.548 & 0.586 & 0.581 & 0.568 & 0.583 & \cellcolor{lightred}{0.522} & 0.570 & 0.565 \\
\bottomrule
\end{tabular}
}
\caption{Performance under base setting and multilingual role-playing setting on MMLU and MMLU-Redux benchmarks. For each row, the \colorbox{lightblue}{best} and \colorbox{lightred}{worst} language results are highlighted.}
\label{table:1}
\end{table*}

\begin{table}[t]
\centering
\resizebox{0.95\linewidth}{!}{
\begin{tabular}{llccc}
\toprule
Model & Benchmark & Improved \textcolor{green}{\textbf{$\uparrow$}}  & Degraded \textbf{\textcolor{red}{$\downarrow$}}  & Unchanged \\
\midrule

\multirow{2}{*}{Qwen3-4B}
& MMLU-Redux & 14 & 12 & 4 \\
& MMLU & 27 & 26 & 4 \\
\cmidrule{1-5}
\multirow{2}{*}{Qwen3-8B}
& MMLU-Redux & 16 & 12 & 2 \\
& MMLU & 28 & 22 & 7 \\
\cmidrule{1-5}
\multirow{2}{*}{Qwen3-14B}
& MMLU-Redux & 19 & 6 & 5 \\
& MMLU & 31 & 17 & 9 \\
\cmidrule{1-5}
\multirow{2}{*}{Qwen3-32B}
& MMLU-Redux & 22 & 4 & 4 \\
& MMLU & 34 & 14 & 9 \\
\cmidrule{1-5}
\multirow{2}{*}{Llama3-8B}
& MMLU-Redux & 8 & 20 & 2 \\
& MMLU & 27 & 23 & 7 \\
\cmidrule{1-5}
\multirow{2}{*}{Mistral-7B}
& MMLU-Redux & 19 & 8 & 3 \\
& MMLU & 25 & 25 & 7 \\

\bottomrule
\end{tabular}
}
\caption{Count statistics of benchmark-internal subjects, comparing model performance under the \textbf{Role-playing} prompt setting and the \textbf{Base} setting. Improved \textcolor{green}{\textbf{$\uparrow$}}, Degraded \textcolor{red}{\textbf{$\downarrow$}}, and Unchanged respectively represent the number of subjects where the role-playing setting outperforms, underperforms, and yields identical results to the base setting.}
\label{table:2}
\end{table}

\subsection{Mechanistic Hypothesis}
The divergent outcomes observed across different models, subjects, and languages suggest that the impact of role-playing is not superficial, but is deeply rooted in how models internalize assigned identities. Drawing inspiration from the meta-cognition concept \cite{ref28}, we propose a mechanistic hypothesis centered on \textbf{persona-related cognitive alignment}: the efficacy of role-playing hinges on whether the LLM accurately conceptualizes the designated persona and its corresponding knowledge domain. Correct conceptualization triggers performance gains, while cognitive deviations hinder them.

\subsection{Validation of the Proposed Hypothesis}
\label{section:Validation}
This premise is initially supported by the performance of Qwen3 across various scales in Tables \ref{table:1} and \ref{table:2}, where larger parameter sizes naturally expand the model's epistemic capacity, leading to better persona comprehension. To rigorous evaluate this hypothesis, we provide further empirical justifications through three analytical lenses: persona informational richness divergence, internal entropy evolution, and thought-space deflection.

\textbf{Validation 1.} To verify that an LLM's cognitive deviation regarding persona information directly dictates its performance gains in role-playing tasks, we conduct a controlled experiment using Qwen3-4B, the weaker model in the Qwen3 series. Specifically, we target subjects where role-playing prompts deliver subpar gains on the MMLU-Redux benchmark. Our experimental paradigm proceeds by systematically injecting role-playing prompts with persona specifications varying in their degree of informational richness. By evaluating how the completeness of persona profiles affects task performance, we aim to establish a clear causal link between the model's cognitive deviations and its subsequent performance fluctuations. 

To eliminate experimental interference from text styling and linguistic expressions, all supplementary persona information is standardized via GPT-5.5 generation, maintaining informational richness as the single controlled variable. We construct three comparative prompt tiers: the \colorbox{yellow!20}{Simple}, \colorbox{orange!20}{Moderate}, and \colorbox{violet!12}{Rich} versions (Appendix \ref{section：Verification Experiment Detail Supplement}).

\begin{table}[htbp]
  \centering
  \resizebox{0.95\linewidth}{!}{
  \begin{tabular}{lccc}
  
    \hline
    Subject & Simple & Moderate & Rich \\
    \hline
    anatomy & $-9\%$ & $-4\%$ & $-4\%$ \\
    high\_school\_geography & $-6\%$ & $-3\%$ & $-1\%$ \\
    college\_physics & $-7\%$ & $-4\%$ & $0\%$ \\
    business\_ethics & $-5\%$ & $-1\%$ & $+2\%$ \\
    \cellcolor{gray!30}{human\_aging} & \cellcolor{gray!30}{$-4\%$} & \cellcolor{gray!30}{$0\%$} & \cellcolor{gray!30}{$-3\%$} \\
    \cellcolor{gray!30}{philosophy} & \cellcolor{gray!30}{$-4\%$} & \cellcolor{gray!30}{$0\%$} & \cellcolor{gray!30}{$0\%$} \\
    professional\_accounting & $-4\%$ & $-1\%$ & $+1\%$ \\
    \hline
  \end{tabular}}
   \caption{Performance deviation of Qwen3-4B under incremental persona information richness settings, where values denote relative performance changes of role-playing prompts compared to the base prompt.}
  \label{table:3}
\end{table}

As demonstrated by the empirical results in Table \ref{table:3}, although performance improvements are observed in most scenarios, role-playing configurations still underperform the baseline across several specific subjects. This phenomenon stems from the model's restricted epistemic capacity. For a model with such constrained expertise, the injection of supplementary persona information merely serves as a partial mitigation strategy rather than a radical remedy for its underlying knowledge gaps. For instance, in \textit{anatomy} and \textit{college\_physics}, expanding the prompt from the Simple Version to the Rich Version either merely mitigates the performance deficit (from $-9\%$ to $-4\%$) or at best brings it back to the baseline ($0\%$), failing to yield any net positive gains over the baseline. 

Crucially, in certain instances, inflating the prompt with excessive contextual details even introduces cognitive noise that actively degrades performance. A prime example is \textit{human\_aging}, where the performance gap stands at $-4\%$ in the Simple Version, recovers to $0\%$ in the Moderate Version, but precipitously drops to $-3\%$ in the Rich Version. This performance degradation arises from excessively redundant persona descriptions, which not only fail to cover the essential knowledge needed to answer targeted questions but also exacerbate cognitive biases, disrupting the model’s judgment.
 
\begin{figure}[htp]   
	\centering
 \includegraphics[width=1\linewidth]{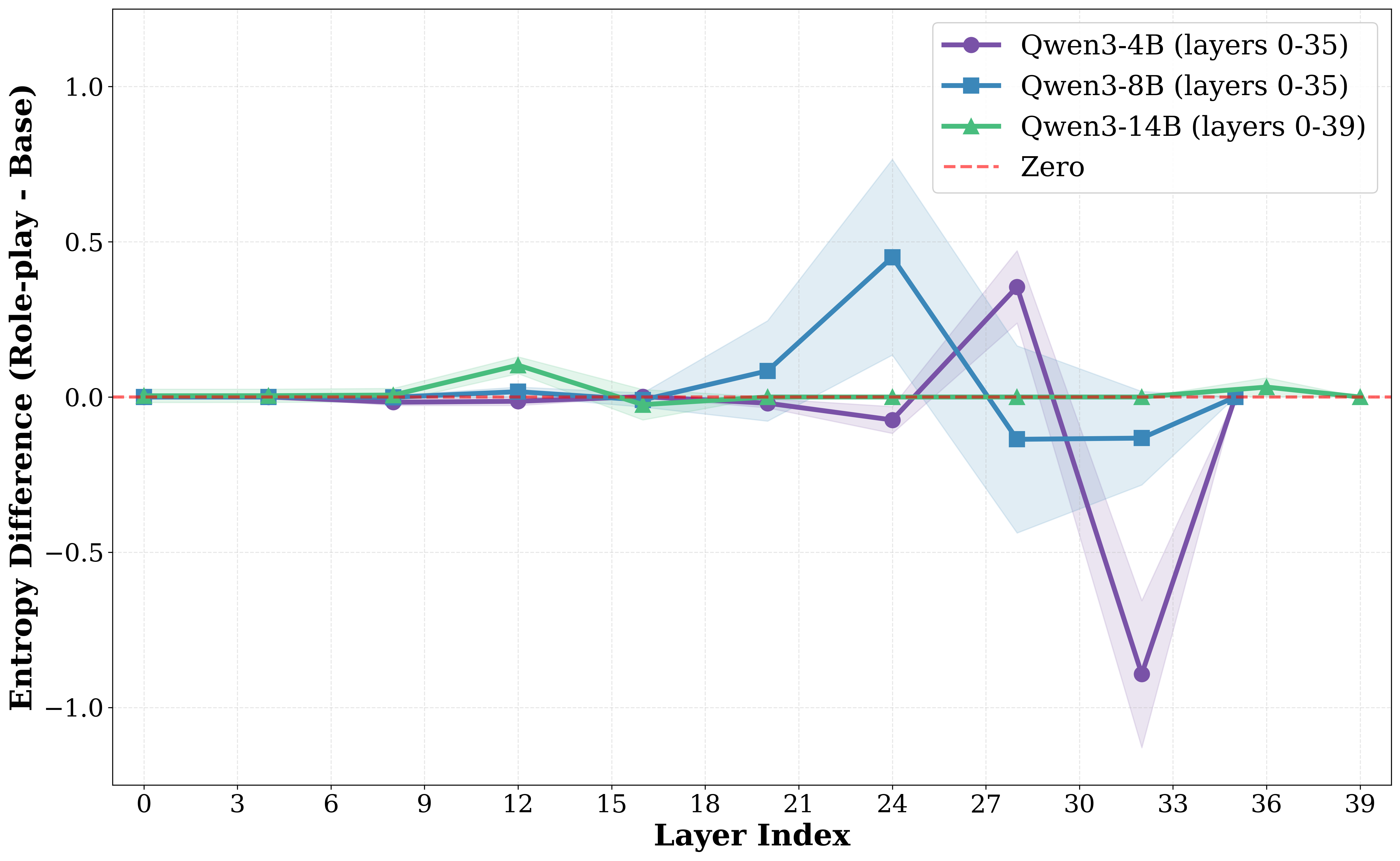}
	\caption{Layer-wise entropy difference across transformer layers for Qwen3-4B/8B/14B on the MMLU-Redux benchmark. Entropy is computed from the final-token hidden state at each layer after projection through the language model head, measuring the model's predictive uncertainty for the next token. Positive values indicate that role-playing prompts increase uncertainty compared to base setting.
    }
	\label{Figure2}
\end{figure}

\textbf{Validation 2.}
Our second verification lens formalizes the logical chain: \textbf{\textit{cognitive deviation $\rightarrow$ representational ambiguity $\rightarrow$ internal entropy evolution}}. Without precise persona conceptualization, the model suffers inference-time cognitive confusion. It cannot efficiently activate relevant knowledge weights, raising uncertainty in intermediate representations; this structural mismatch produces volatile entropy shifts \cite{ref29}. We thus calculate layer-wise entropy divergence between role-play and baseline settings, with results plotted in Figure \ref{Figure2}.

Entropy divergence stays close to zero in the first 16 shallow feature-extraction layers. Sharp separation arises in layers 20–35, the deep semantic abstraction and decision stage. The stronger Qwen3-14B resolves representational ambiguity and clearly partitions persona knowledge, maintaining near-zero, steady entropy divergence across layers. In contrast, weaker Qwen3-4B/8B exhibit severe cognitive deviation and confusion. Their hidden states suffer strong inter-layer conflicts, driving wild entropy fluctuations in mid-to-late layers—the network segments responsible for high-level semantic integration and persona alignment.

\textbf{Validation 3.}
Our third analytical lens examines thought-space deflection to substantiate our hypothesis. Specifically, we select three representative subjects from Qwen3-8B characterized by the maximum positive gain, maximum negative gain, and neutral performance under role-playing setups. We analyze their internal hidden states by calculating the $L2$ Euclidean norms and applying PCA \cite{ref13} dimensionality reduction to map the representations into a 2D space. 

\begin{figure}[htp]   
	\centering
 \includegraphics[width=1\linewidth]{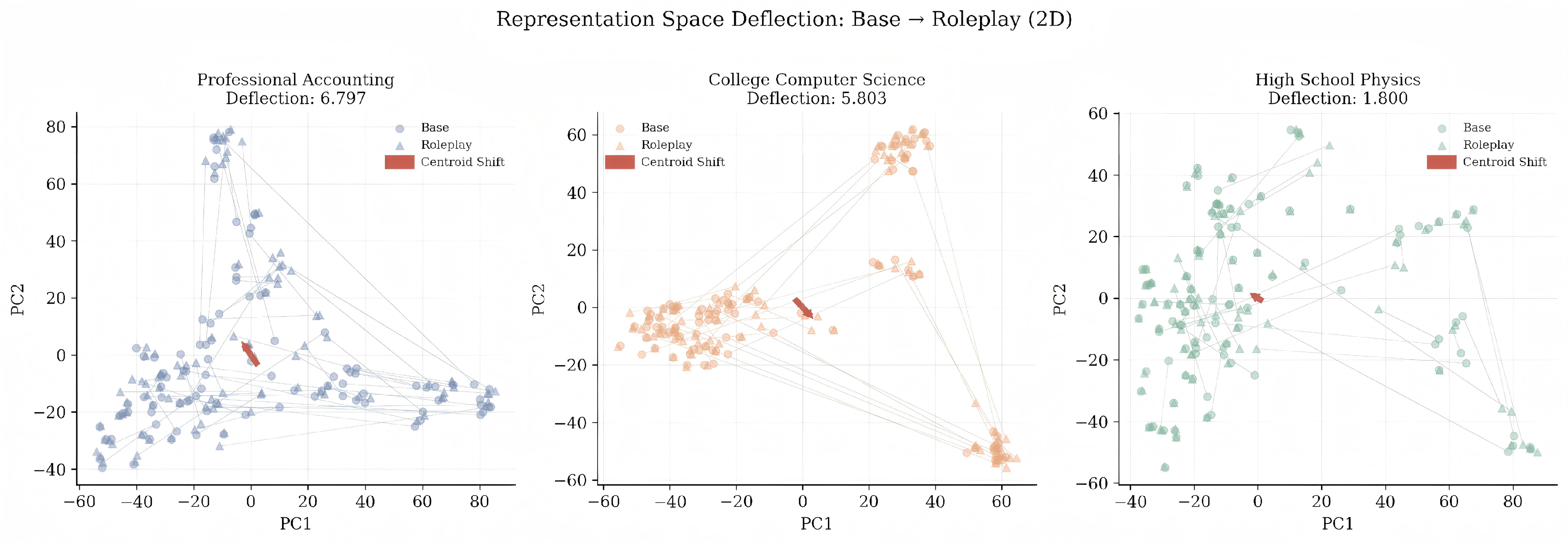}
	\caption{Comparison of hidden representation space deflection derived from layer-averaged states between base and role-playing prompting on multiple MMLU-Redux subjects.}
	\label{Figure3}
\end{figure}

As depicted in Figure \ref {Figure3}, both the maximum-positive and maximum-negative gain subjects exhibit conspicuous deflection amplitudes, but their shifting orientations are distinctly divergent. In contrast, the subject with no performance variance demonstrates negligible spatial shift. This provides explicit evidence that role-playing persona prompts trigger geometric deflections within the model's inner representation space. Crucially, while the performance variance dictates the magnitude of this spatial deflection, the specific direction of the deflection is rooted in the model's cognitive alignment or deviation regarding the persona, which fundamentally determines whether role-playing enhances or degrades model utility.

\section{Methodology}

Our preliminary study demonstrates that the effectiveness of role-playing prompting is highly sensitive to prompt formulation, particularly the linguistic expression of expert profiles. This observation suggests that different linguistic realizations of semantically equivalent role prompts may induce diverse internal representations within LLM. However, manually identifying an optimal prompt formulation remains challenging, as users typically lack prior knowledge about which linguistic representation can better align with the expertise required for a given task.

\begin{figure}[t]   
	\centering
 \includegraphics[width=1\linewidth]{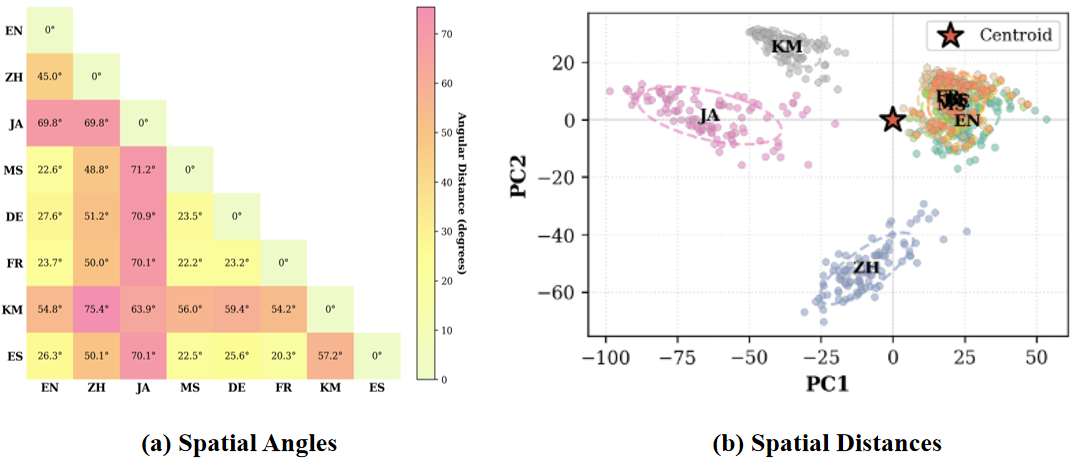}
	\caption{Spatial geometric analysis of hidden representations elicited by Qwen3-14B. (a) Pairwise angular distances between mean representation centroids induced by role-playing prompts in eight languages. (b) 2D PCA visualization of last-layer hidden states.}
	\label{Figure4}
\end{figure}

\begin{figure}[t]   
	\centering
 \includegraphics[width=1\linewidth]{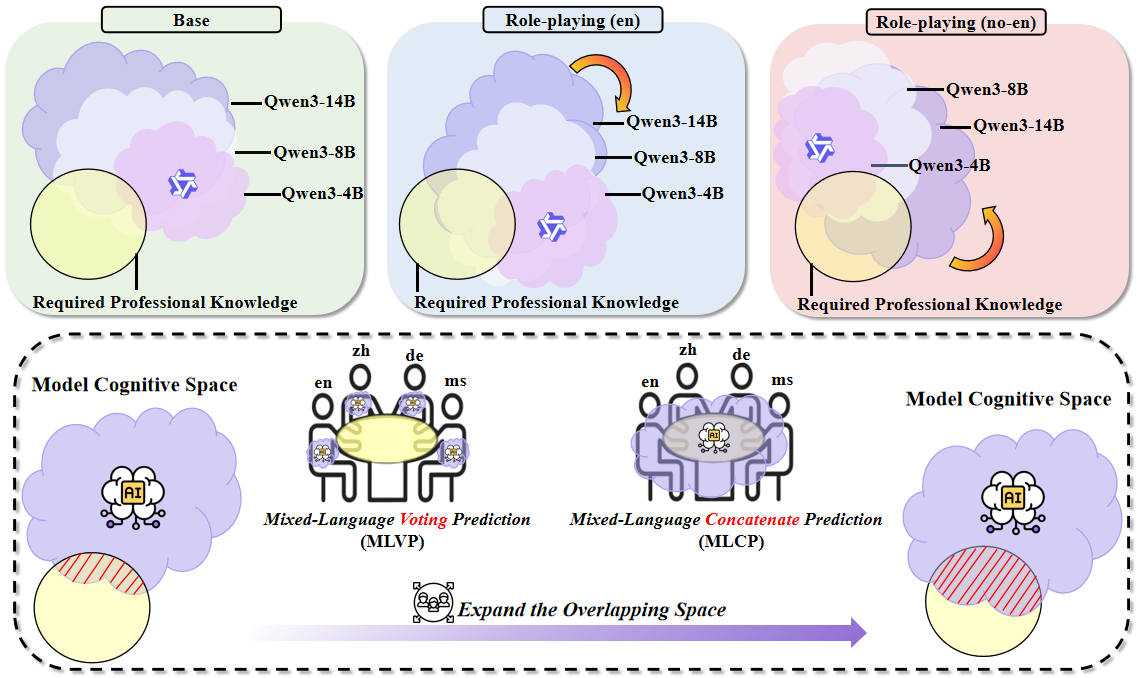}
	\caption{Illustration of multilingual role-playing prompt composition. Different languages induce diverse representation patterns, and their composition provides a broader inference context.}
	\label{Figure5}
\end{figure}

To investigate this phenomenon, we analyze the geometric characteristics of hidden representations elicited by role-playing prompts across different languages. Specifically, we measure the angular distances and positional shifts among hidden representation spaces induced by multilingual role prompts, as shown in Figure~\ref{Figure4}. The results reveal that different languages introduce distinct representation deviations within the LLM's latent space, suggesting that multilingual formulations provide diverse representation patterns for the same semantic instruction. These geometric observations motivate us to explore whether language-induced representation diversity can be leveraged during inference. As illustrated in Figure~\ref{Figure5}, different linguistic realizations of the same role prompt correspond to different regions in the representation space. Instead of selecting a single prompt formulation, composing multiple language-specific role prompts provides the model with complementary representation cues within a unified context.

Inspired by self-consistency mechanisms \cite{ref14}, multilingual prompting for generation diversity \cite{ref31}, and prompt repetition strategies \cite{ref27}, we propose \textbf{Mixed-Language Concatenate Prediction (MLCP)}, an inference-time strategy that composes semantically equivalent role-playing prompts expressed in different languages before generation. Different from conventional self-consistency approaches that independently sample multiple responses and aggregate predictions, MLCP performs multilingual prompt composition within a single inference process, enabling the model to jointly utilize diverse linguistic role cues.

Given a role-playing prompt configuration $\mathcal{P}_{\text{rp}}$ and a multilingual language set $\mathcal{L}=\{l_1,l_2,...,l_K\}$, MLCP constructs multiple language-specific role prompts by preserving the semantic meaning of the original expert profile while varying its linguistic realization. Specifically, each language-specific prompt is generated as:
\begin{equation}
\mathcal{P}_{i}
=
\phi(\mathcal{P}_{\text{rp}}, l_i),
\quad l_i \in \mathcal{L}
\end{equation}
where $\phi(\cdot)$ denotes the language transformation function and $l_i$ represents the language used to express the role instruction. The resulting prompts maintain identical task semantics while introducing diverse linguistic formulations. MLCP then composes these semantically equivalent multilingual role prompts into a unified context:
\begin{equation}
\mathcal{P}_{\text{MLCP}}
=
[\mathcal{P}_{1};\mathcal{P}_{2};...;\mathcal{P}_{K}]
\end{equation}

\begin{table*}[t]
\centering
\resizebox{1\linewidth}{!}{
\begin{tabular}{l cccccc|cccccc}
\toprule
\multirow{2}{*}{Model} & \multicolumn{6}{c|}{MMLU-Redux} & \multicolumn{6}{c}{MMLU} \\
\cmidrule(lr){2-7} \cmidrule(lr){8-13}
 & Base & Role-playing & MLVP & SLVP-best & MLCP & SLCP-best & Base & Role-playing & MLVP & SLVP-best & MLCP & SLCP-best \\
\midrule
Qwen3-4B   & 0.756 & 0.754 & 0.769 & 0.764 & 0.779 & \textbf{0.782} & 0.756 & 0.754 & 0.798 & 0.773 & \textbf{0.807} & 0.779 \\
Qwen3-8B   & 0.764 & 0.767 & 0.790 & 0.779 & \textbf{0.822} & 0.819 & 0.782 & 0.782 & 0.821 & 0.801 & \textbf{0.847} & 0.815 \\
Qwen3-14B  & 0.798 & 0.810 & 0.821 & 0.825 & \textbf{0.843} & 0.841 & 0.822 & 0.824 & 0.854 & 0.836 & \textbf{0.866} & 0.844 \\
Qwen3-32B & 0.834 & 0.852 & 0.850 & 0.849 & \textbf{0.868} & 0.865 & 0.841 & 0.856 & 0.872 & 0.856 & \textbf{0.884} & 0.862 \\
Llama3-8B  & 0.539 & 0.524 & 0.525 & 0.529 & \textbf{0.545} & 0.522 & 0.611 & 0.618 & 0.619 & 0.614 & \textbf{0.621} & 0.594 \\
Mistral-7B & 0.458 & 0.467 & 0.475 & 0.473 & \textbf{0.481} & 0.471 & 0.588 & 0.589 & \textbf{0.597} & 0.591 & 0.592 & 0.576 \\
\bottomrule
\end{tabular}
}
\caption{Performance comparison of different models and methods on MMLU-Redux and MMLU. SLVP-best and SLCP-best represent the best-performing language in \textit{Single-Language Voting Prediction} and \textit{Single-Language Concatenate Prediction}, respectively. \textbf{Bold} indicates the best-performing setting per model and benchmark.}
\label{table:4}
\end{table*}

Unlike conventional self-consistency strategies that independently sample multiple responses and aggregate predictions, MLCP performs multilingual prompt composition before inference, allowing the model to jointly process diverse linguistic role cues within a single context. By integrating multiple linguistic realizations of the same expert profile, MLCP aims to exploit complementary representation patterns induced by different languages, thereby improving the robustness and generalization of role-playing prompting.

To further investigate the underlying factors contributing to the effectiveness of multilingual prompt composition, we design three controlled alignment strategies: (a) \textbf{Mixed-Language Voting Prediction (MLVP)}: independently generates responses from semantically equivalent prompts expressed in multiple languages and obtains the final prediction through majority voting; (b) \textbf{Single-Language Voting Prediction (SLVP)}: repeatedly samples responses using a single-language role prompt and aggregates predictions through majority voting; (c) \textbf{Single-Language Concatenate Prediction (SLCP)}: repeat and concatenate identical monolingual role prompts to disentangle prompt effects from multilingual representational diversity.

\section{Experimental Results and Analysis}

\subsection{Main Results}

Table \ref{table:4} summarizes the performance of the proposed variant across multiple models. Among all variants, MLCP achieves the best overall performance and outperforms the other three variants on most benchmarks and model. Meanwhile, it yields significant performance gains compared with the unoptimized role-playing setup.

In addition, we present the subject count comparison between the role-playing and MLCP setting in Table \ref{table:5}. It can be observed that, under the MLCP strategy, the number of subjects yielding positive gains increases markedly, while the number of subjects showing negative gains decreases substantially. This trend suggests that MLCP not only enhances overall prediction performance but, more critically, effectively alleviates the cognitive bias commonly observed in role-playing scenarios. Specifically, grounded in the cognitive deviation hypothesis validated by our preliminary study, different linguistic phrasings induce distinct spatial deflections within the model's epistemic representation space. By concatenating identical-semantic multilingual prompts, MLCP integrates multiple heterogeneous cognitive spaces, achieving broader topological alignment and overlap with the prerequisite domain expertise required to correctly address the target task. This multi-angle persona alignment effectively counteracts underlying cognitive deviations and expands the intersection between the model's epistemic boundaries and the necessary expert knowledge, ultimately maximizing role-playing performance gains.
\begin{table}[htp]
\centering
\resizebox{1\linewidth}{!}{
\begin{tabular}{llccc}
\toprule
Model & Benchmark & Improved \textcolor{green}{\textbf{$\uparrow$}}  & Degraded \textbf{\textcolor{red}{$\downarrow$}}  & Unchanged  \\
\midrule

\multirow{2}{*}{Qwen3-4B}
& MMLU-Redux & 14$\boldsymbol{\rightarrow}$19 & 12$\boldsymbol{\rightarrow}$9 & 4$\boldsymbol{\rightarrow}$2 \\
& MMLU & 27$\boldsymbol{\rightarrow}$42 & 26$\boldsymbol{\rightarrow}$10 & 4$\boldsymbol{\rightarrow}$5 \\
\cmidrule{1-5}
\multirow{2}{*}{Qwen3-8B}
& MMLU-Redux & 16$\boldsymbol{\rightarrow}$26 & 12$\boldsymbol{\rightarrow}$2 & 2$\boldsymbol{\rightarrow}$2 \\
& MMLU & 28$\boldsymbol{\rightarrow}$49 & 22$\boldsymbol{\rightarrow}$5 & 7$\boldsymbol{\rightarrow}$3 \\
\cmidrule{1-5}
\multirow{2}{*}{Qwen3-14B}
& MMLU-Redux & 19$\boldsymbol{\rightarrow}$22 & 6$\boldsymbol{\rightarrow}$4 & 5$\boldsymbol{\rightarrow}$4 \\
& MMLU & 31$\boldsymbol{\rightarrow}$39 & 17$\boldsymbol{\rightarrow}$13 & 9$\boldsymbol{\rightarrow}$5 \\
\cmidrule{1-5}
\multirow{2}{*}{Qwen3-32B}
& MMLU-Redux & 22$\boldsymbol{\rightarrow}$26 & 4$\boldsymbol{\rightarrow}$3 & 4$\boldsymbol{\rightarrow}$1 \\
& MMLU & 34$\boldsymbol{\rightarrow}$46 & 14$\boldsymbol{\rightarrow}$8 & 9$\boldsymbol{\rightarrow}$3 \\
\cmidrule{1-5}
\multirow{2}{*}{Llama3-8B}
& MMLU-Redux & 8$\boldsymbol{\rightarrow}$10 & 20$\boldsymbol{\rightarrow}$17 & 2$\boldsymbol{\rightarrow}$3 \\
& MMLU & 27$\boldsymbol{\rightarrow}$27 & 23$\boldsymbol{\rightarrow}$21 & 7$\boldsymbol{\rightarrow}$9 \\
\cmidrule{1-5}
\multirow{2}{*}{Mistral-7B}
& MMLU-Redux & 19$\boldsymbol{\rightarrow}$21 & 8$\boldsymbol{\rightarrow}$7 & 3$\boldsymbol{\rightarrow}$2 \\
& MMLU & 25$\boldsymbol{\rightarrow}$26 & 25$\boldsymbol{\rightarrow}$23 & 7$\boldsymbol{\rightarrow}$8 \\

\bottomrule
\end{tabular}
}
\caption{Statistics on the number of benchmark subjects with performance differences.
Each entry follows the format $A \rightarrow B$: $A$ counts subjects where role-playing outperforms/degrades/equals the base setting,
while $B$ counts subjects where MLCP outperforms/degrades/equals the base setting.}

\label{table:5}
\end{table}

\subsection{Quantitative Analysis of Mixed Languages}

\begin{figure}[htp]   
	\centering
 \includegraphics[width=1\linewidth]{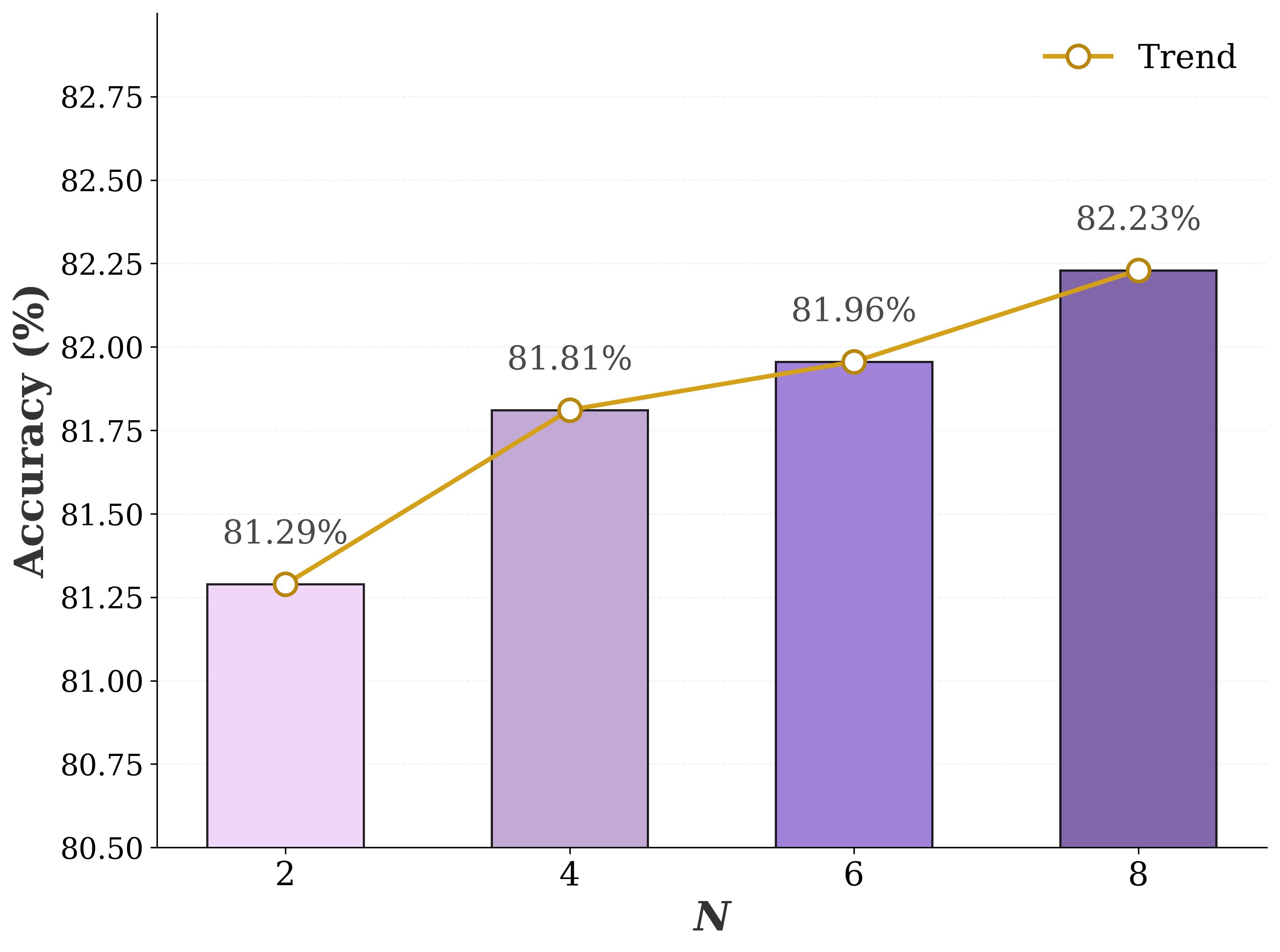}
	\caption{Performance comparison of Qwen3-8B on MLCP variant across 2–8 languages. The \textit{N} denotes the number of mixed languages.}
	\label{Figure6}
\end{figure}

This section aims to quantitatively investigate the impact of the number of mixed languages (\textit{N}) incorporated into the input on the final performance of the MLCP variant. To eliminate biases introduced by specific language combinations, for each given \textit{N} (\textit{N}=2, 4, 6, 8), we randomly selected different language combinations from the overall language pool and conducted multiple independent trials, computing the average accuracy across all runs. The experimental results are presented in Figure \ref{Figure6}. A clear and consistent monotonic upward trend can be observed: as the number of mixed languages increases from 2 to 8, the average accuracy of the model rises substantially, indicating that the model benefits from exposure to a more diverse set of languages.

This finding strongly corroborates the original design rationale of the MLCP variant articulated in our methodology. As theorized, variations in linguistic phrasing can be conceptualized as distinct geometric deflections within the model's cognitive space, and the degree of overlap between these deflected spaces and the prerequisite domain knowledge directly determines task performance. By concatenating semantically equivalent prompts across an increasing number of languages, MLCP effectively constructs a more comprehensive and multi-angle persona alignment, expanding the intersection between the model's cognitive boundaries and the required expert knowledge. The monotonic performance gain with increasing \textit{N}, as demonstrated in Figure \ref{Figure6}, directly provides empirical support: a broader spectrum of linguistic phrasings induces richer spatial deflections, which in turn yields more thorough coverage of the target knowledge space. This quantitative analysis thus provides solid empirical evidence for the efficacy of the MLCP framework, confirming that leveraging multilingual diversity constitutes a viable path toward maximizing role-playing performance gains.

\section{Related Work}
Role-playing prompting has emerged as a lightweight, cost-effective paradigm for optimizing LLMs. Without expensive fine-tuning, it regulates outputs, mitigates hallucinations, and enhances task adaptability through persona alignment and behavioral constraints \cite{ref15,ref16}. Consequently, it has been widely adopted across natural language generation, intelligent dialogue, and logical reasoning. It also plays a pivotal role in frontiers such as multi-agent collaboration, LLM harness, and skills design, serving as a core mechanism to orchestrate agent division of labor and activate specialized capabilities within general-purpose foundation models \cite{ref17,ref18,ref19,ref20,ref21}.

Despite its undeniable practical utility, current research on role-playing prompting is constrained by pronounced contextual limitations. Most existing studies focus heavily on isolated scenarios including interactive dialogue \cite{ref22,ref23}, creative writing \cite{ref24}, or mathematical logical reasoning \cite{ref4,ref25,ref26}, and merely validate the efficacy or necessity of role-playing under localized task settings. Consequently, these findings remain fragmented empirical observations tailored to specific applications, lacking a generalized and holistic perspective \cite{ref32}.


\section{Conclusion}
This paper challenges the prevailing belief that role-playing universally enhances LLM performance. Through comprehensive experiments, we demonstrate that its efficacy correlates with multiple factors, including model capability, target subject, and prompt language. Specifically, stronger models benefit significantly, whereas weaker ones often suffer degradation; furthermore, no single domain yields equal performance gains across all models through role-playing, and high-resource languages do not guarantee optimal results.

Mechanistically, our analysis indicates that role-playing functions as a conditional knowledge activation mechanism. Performance correlates with the model's cognitive alignment with the target persona; deviations manifest as representational ambiguity, entropy fluctuations, and thought-space deflection, ultimately undermining results. Building on this, we propose multilingual alignment strategies, particularly MLCP, to effectively expand the intersection between model cognition and domain expertise. Our work provides a rigorous foundation for optimizing role-playing, emphasizing context-aware, mechanism-driven design over simplistic empirical adoption.


\clearpage
\section*{Limitation}

Our evaluation covers six open-weight models from three families, none above 32B, on two multiple-choice knowledge QA benchmarks, so the correlation between role-playing gains and model capability is established mainly within the Qwen3 series and should be read as a well-supported hypothesis rather than a universal law; extending it to larger models, other families, and open-ended or interactive settings is a natural next step. The persona-related cognitive alignment hypothesis is an interpretive framework supported by three converging lines of evidence—persona-richness manipulation, layer-wise entropy divergence, and thought-space deflection—each with its own confounds, and our mechanistic analyses concentrate on the Qwen3 family, which provides a controlled setting for isolating scale effects, so the evidence is consistent with the hypothesis rather than a direct proof. Our multilingual comparison varies the persona-instruction language while keeping questions in English, so the observed cross-language variation reflects both persona-related and general cross-lingual effects. Finally, MLCP is deliberately simple and training-free, and its contribution lies in showing that pre-inference composition of semantically equivalent role prompts yields stable gains over vanilla role-playing; because it lengthens the input, the accuracy-cost trade-off and adaptive language selection remain open questions.

\section*{Ethics Statement}

We confirm that this study follows the Ethics Policy. 
All data and models used are publicly available and contain no private user information.

\paragraph{Data and Model Bias}
Our experiments use only publicly available benchmarks (MMLU and MMLU-Redux) and open-weight
language models. We do not collect or use any private user data. These benchmarks and models are
widely adopted in the research community and contain no known systematic biases that would affect
the validity of our conclusions. Our method, MLCP, only composes multilingual role-playing prompts
at inference time and does not rely on or amplify any sensitive attributes.

\paragraph{Intended Use}
MLCP is designed as a lightweight, training-free inference strategy to improve the robustness of
role-playing prompting in knowledge QA. It is intended for research and practical use in scenarios
where users wish to obtain more stable gains from persona-based prompts without additional
training, and it should not be used to generate misleading or impersonating content.

\bibliography{custom}

\appendix

\section{Frequently Asked Questions (FAQs)}
\subsection{Code and Data Availability}

To facilitate future research and allow independent verification of our results, all code and
data associated with MLCP are released via an anonymous repository:
\textcolor{blue}{\url{https://anonymous.4open.science/r/RethinkingRolePlay}}.
This repository provides a comprehensive experimental framework. Specifically, it details our
exact methodology, including the multilingual role-prompt construction, the MLCP concatenation
procedure, and the implementation of all comparison variants (MLVP, SLVP, and SLCP). Coupled
with the release of all processed evaluation splits, standardized multilingual prompt templates,
answer-extraction scripts, and full evaluation scripts, this repository grants researchers
unrestricted access to every component required to reproduce our work.

\subsection{What Are the Advantages of Our Work?}
The core vulnerability of existing research lies in the absence of a systematic, generalized analysis of the underlying mechanisms driving role-playing prompting. Crucial foundational questions remain unanswered within the academic community: Is role-playing prompting a universal enhancement mechanism capable of comprehensively boosting model performance, or is it merely a contextual adaptation strategy heavily dependent on task-specific attributes? Present research lacks cross-task and cross-domain comparative validation. As a result, the precise functional boundaries, optimal adaptive conditions, and failure modes of this strategy remain obscured, leaving the performance variance across different tasks unexplained.

In summary, current exploration remains confined to single-scenario validation and localized optimization, lacking a systematic inquiry into the intrinsic nature, universal validity, and task dependency of role-playing prompts. This gap severely hinders the generalized deployment and iterative optimization of this paradigm. To address these research desiderata, this paper systematically investigates the underlying mechanisms of role-playing prompting. Through comprehensive multilingual and multi-domain comparative experiments, we delineate its universal empowerment capabilities and contextual adaptation boundaries, thereby filling a critical void in prior work.

\subsection{Is MLCP Merely an Incremental Extension of Existing Prompt Composition Strategies?}

At the mechanism level, yes: MLCP composes semantically equivalent multilingual role prompts and is inspired by self-consistency, multilingual prompting, and prompt repetition, so we do not claim a new inference method. This is deliberately an analysis-heavy, method-light paper: before proposing MLCP, we conduct a multi-model, cross-domain, multilingual evaluation of role-play prompting on MMLU and MMLU-Redux, propose the persona-related cognitive alignment hypothesis, and validate it through persona-richness ablation, layer-wise entropy divergence, and thought-space deflection, as well as a cross-lingual representation analysis showing that different languages induce distinct geometric deflections for the same persona; MLCP is the minimal strategy derived directly from these findings. That MLCP is simple is thus intentional, and it is not reducible to repetition: SLCP—which repeats and concatenates monolingual prompts, matching MLCP in length, repetition, and input diversity—is outperformed by MLCP in most settings (Table \ref{table:4}), and MLCP improves monotonically as the number of languages grows from 2 to 8 (Figure \ref{Figure6}), which a pure repetition account would not predict.

\subsection{Does Persona Richness Simply Provide Better Contextual Cues?}

A natural concern is that enriching the persona also increases prompt length and adds
task-relevant concepts, so gains may come from richer contextual cues rather than from better
persona conceptualization. We define persona conceptualization as the extent to which the
induced representation of the assigned expert overlaps with the knowledge required by the
question; this is not equivalent to prompt length or injected information. We therefore use
richness only as a perturbation of the persona representation, not as a measurement of
conceptualization. If gains were driven solely by additional cues, richer prompts should
improve performance monotonically. Our results contradict this: on \texttt{human\_aging},
performance moves from $-4\%$ (Simple) to $0\%$ (Moderate) and then to $-3\%$ (Rich), and on
\texttt{anatomy} and \texttt{college\_physics} richness only mitigates the deficit without
yielding net gains. Such non-monotonic, subject-dependent behavior is hard to explain by
generic cues alone. We acknowledge that richness co-varies with length and specificity, and
thus treat this as one of three converging signals rather than a standalone proof; isolating
persona alignment via length-controlled or domain-mismatched variants is left for future work.

\subsection{Does the Multilingual Manipulation Isolate Persona-Related Cognition?}

We acknowledge that our multilingual design keeps the benchmark questions in English and
varies only the language of the persona instruction, so the manipulation also introduces
code-switching and engages the model's uneven cross-lingual proficiency; the large degradation
under low-resource persona languages (e.g., Llama3-8B drops from $0.524$ under English to
$0.249$ under Khmer on MMLU-Redux) therefore cannot be attributed to persona misalignment
alone, and we do not treat language-based performance differences as direct evidence of
persona-specific alignment. This confound, however, does not fully explain our results. Our
SLCP control repeats and concatenates identical monolingual role prompts, matching MLCP in length, repetition, and input diversity while holding linguistic variety fixed, yet MLCP still outperforms it in the vast majority of cases (Table \ref{table:4}), so the gains cannot be attributed to tokenization, language familiarity, or generic prompt-diversity effects alone.

Moreover, if cross-language
variation were driven purely by familiarity or tokenization, high-resource languages should
dominate, whereas for Qwen3 the Chinese role prompt does not yield the best performance and the
optimal language varies across models and subjects, which is more consistent with different
linguistic realizations inducing different internal representations of the same persona. 

\subsection{Is MLCP Compared Against the Strongest Persona Baselines?}

The ``Role-playing'' column in Table \ref{table:4} reports the English persona prompt, which is the
natural default for English benchmark questions and the setting implicitly assumed by most
prior work; we use it as the primary reference rather than the best-performing language per
model and benchmark, since selecting the latter would require test-set tuning and would not
reflect realistic usage. We nonetheless report the best single-language results (SLVP-best and
SLCP-best) and note that MLCP does not exceed the best observed single language in every
setting (e.g., Mistral-7B on MMLU-Redux: $0.494$ vs.\ $0.481$); MLCP's advantage is its
robustness across models and languages rather than uniform dominance over every per-language
optimum.

\section{Preliminary Study Setup Details}

\subsection{Datasets}
We evaluate these models on two multiple-choice benchmarks characterized by broad disciplinary coverage:
\begin{itemize}
    \item \textbf{MMLU} \cite{ref8}, a comprehensive benchmark comprising 57 fine-grained subjects that span a wide range of domains, including the humanities, social sciences, and STEM. It is designed to evaluate models' broad knowledge and reasoning capabilities across diverse academic disciplines, covering areas such as history, law, mathematics, physics, and computer science.
    
    \item \textbf{MMLU-Redux} \cite{ref9}, a refined variant of MMLU comprising 30 subjects, explicitly curated to eliminate potential label noise and ambiguous items. By removing noisy or problematic questions, MMLU-Redux provides a cleaner and more reliable evaluation set, making it better suited for accurately assessing model performance and reducing evaluation bias.
\end{itemize}

\subsection{Evaluation Prompts}
\begin{figure}[t]   
	\centering
 \includegraphics[width=1\linewidth]{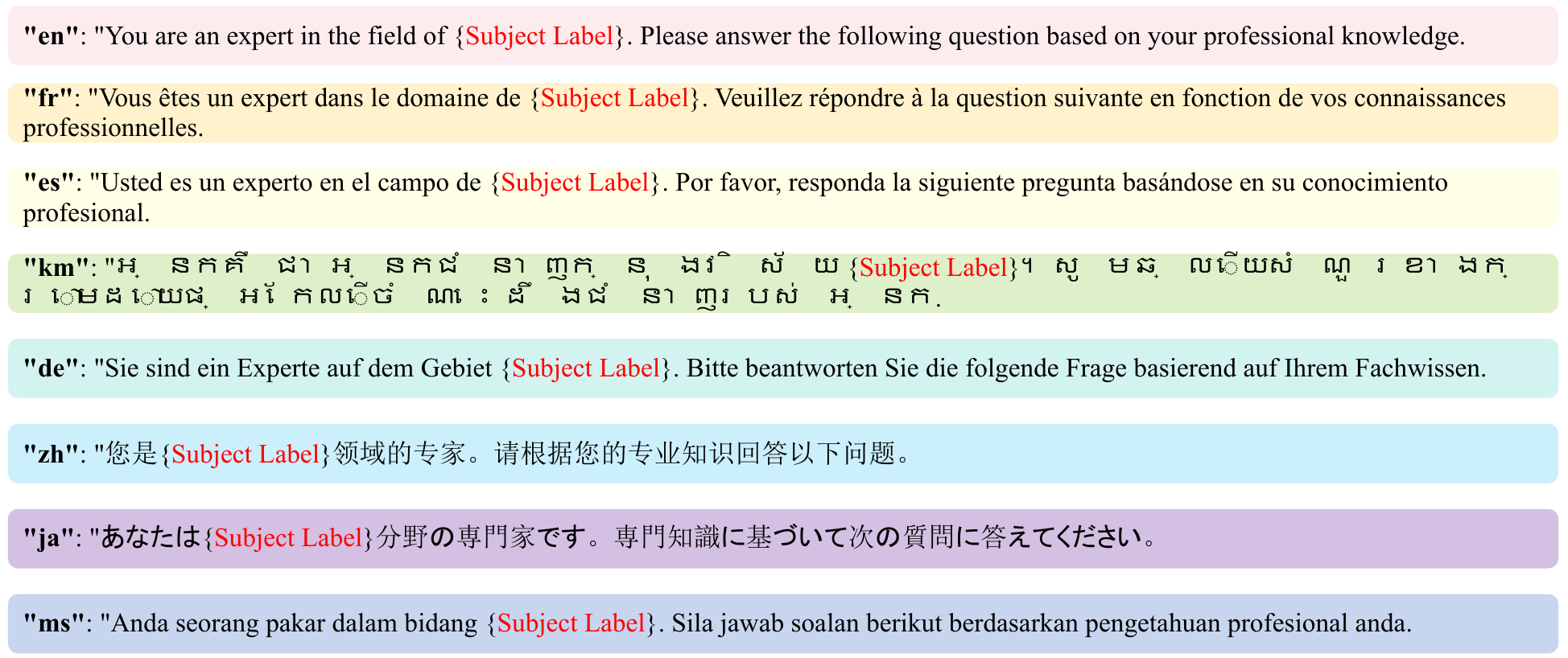}
	\caption{Multilingual role-playing prompts.}
	\label{Figure7}
\end{figure}

As described in Section \ref{section:Study Setup} (Study Setup) of the main paper, we extend role-play evaluation to multilingual scenarios. To simulate real-world interactions where users naturally craft role-playing prompts in their native languages, we extend the role-playing evaluation into a multilingual scenario. We translate the persona instructions into eight distinct languages: English (en), Chinese (zh), Spanish (es), French (fr), Japanese (ja), German (de), Khmer (km), and Malay (ms). Crucially, these prompts maintain strict semantic equivalence, isolating language as the sole independent variable. The specific multilingual prompts are illustrated in Figure \ref{Figure7}.

We apply the template shown in Table~\ref{table:6} to conduct the role-playing tests in the preliminary study.

\begin{table*}[htbp]
\centering
\begin{tcolorbox}[
  colback=gray!10,
  colframe=black!100,
  title=Role-playing Test Template,
  width=\linewidth
]
You are an expert in the field of \texttt{[Subject Label]}. Please answer the following question based on your professional knowledge.

Question: \texttt{[Question]}

A. \texttt{[choice\_0]}

B. \texttt{[choice\_1]}

C. \texttt{[choice\_2]}

D. \texttt{[choice\_3]}

Please output only the letter of the correct answer (A, B, C, or D). Do not output any other text.

Model Output: \texttt{[pred answer]}
\end{tcolorbox}

\caption{Prompt template used for role-playing question answering throughout our experiments.}
\label{table:6}
\end{table*}

\section{Verification Experiment Detail Supplement}
\label{section：Verification Experiment Detail Supplement}
To validate our hypothesis, we test and analyze persona informational richness divergence in Validation 1 of Section \ref{section:Validation}. Our experiments include seven test subjects with three prompt versions per subject. Relevant details are provided in Tables \ref{table:7}--\ref{table:9}.

\section{Subject Results Supplement}

We supplement the subject-level results for both benchmarks. MMLU-Redux results are reported in Tables \ref{table:10} and \ref{table:11}, covering 30 subjects across Mistral-7B, Llama3-8B, Qwen3-4B, Qwen3-8B, Qwen3-14B and Qwen3-32B. MMLU results are split across Tables~\ref{table:12}--\ref{table:14} due to the larger number of subjects (57 in total).

These tables reveal that role-playing effects are highly conditional. The same model can improve on some subjects and degrade on others, and models of different scales within the same family can even move in opposite directions on the same subject. For example, on MMLU-Redux anatomy, Qwen3-4B drops by 9\% while Qwen3-14B gains 1\%; on \textit{college\_physics}, Qwen3-4B drops by 7\% while Qwen3-8B gains 5\%. No subject shows consistent gains across all models. These observations reinforce the paper's central claim that role-playing is not a universal enhancement mechanism but a context-dependent adaptation strategy shaped jointly by model capability, knowledge domain, and question type.

\section{Performance of Each Language in SLVP and SLCP Variants}

Table \ref{table:15} reports the full language-level results for SLVP and SLCP across eight languages (en, zh, ja, km, ms, de, es, fr) on MMLU-Redux and MMLU. The main text selects only the best-performing language from each variant family as a comparison baseline; this table provides the complete picture.

Two findings stand out. First, performance varies substantially across languages. For Llama3-8B on MMLU-Redux SLVP, English reaches 0.523 while Khmer falls to 0.309, a gap of over 20 points. Even high-resource languages such as English and Chinese do not consistently yield the best results. Second, SLCP generally outperforms SLVP, confirming that prompt concatenation itself contributes gains, but SLCP still lags behind MLCP and fluctuates across languages. The optimal language also shifts depending on model, benchmark, and variant, making it difficult for users to pre-select a single best language. This variability motivates MLCP: by composing semantically equivalent role prompts across multiple languages in one context, the model can exploit complementary representational cues and reduce reliance on any single language choice.

\section{Experimental Hardware \& Computing Environment}

All experiments are implemented on a server running the Ubuntu 22.04.3 LTS operating system, which is configured with an Intel(R) Xeon(R) Platinum 8338C CPU at 2.60 GHz, 1.0 TiB of system memory, and NVIDIA A800-SXM4-80GB GPUs with 81920 MiB video memory for each graphics card.

\section{Usage of AI Assistants}
AI assistants were used only for linguistic refine
ment of the manuscript (clarity of arguments, log
ical coherence, and grammatical correctness) and
for debugging our implementation code. The re
search idea, experimental design, and analysis of
results were performed entirely by the authors, who
assume full responsibility for the scientific con
tent, conclusions, and integrity of this paper, and
confirm compliance with academic ethics and the
absence of plagiarism.

\begin{table*}[htbp]
\centering

\small

\begin{tcolorbox}[
    colback=gray!5!white,
    colframe=gray!50!black,
    coltitle=white,
    fonttitle=\bfseries\small,
    fontupper=\small,
    boxsep=2pt,
    left=3pt, right=3pt, top=2pt, bottom=2pt,
    title=1. Anatomy
]
\renewcommand{\arraystretch}{1.15}
\begin{tabularx}{\linewidth}{@{}p{2.2cm} X@{}}
\toprule
\textbf{Version} & \textbf{Prompt} \\
\midrule
\colorbox{yellow!20}{\textbf{Simple}} &
You are an expert in the field of anatomy. Please answer the following question based on your professional knowledge. \\
\colorbox{orange!20}{\textbf{Moderate}} &
You are a professional expert in human anatomy. You have solid knowledge of major organ systems, including skeleton, muscles, nerves, circulation, respiration, digestion, and reproduction. You are familiar with standard anatomical terms and the structural relationships of tissues and organs. Please select the correct answer based on authoritative anatomical knowledge. \\
\colorbox{violet!12}{\textbf{Rich}} &
You are a professional expert in human anatomy. You have comprehensive knowledge of human body structures, covering major organ systems such as skeleton, muscles, nerves, blood circulation, respiration, digestion, urinary and reproductive systems. You are familiar with standard anatomical terms, the positional relationship, structural features and physiological functions of various tissues and organs. You can tell apart common structural features and typical anatomical knowledge points. Please analyze the question carefully and make your choice strictly based on authoritative anatomical knowledge. \\
\bottomrule
\end{tabularx}
\end{tcolorbox}

\vspace{1mm}

\begin{tcolorbox}[
    colback=gray!5!white,
    colframe=gray!50!black,
    coltitle=white,
    fonttitle=\bfseries\small,
    fontupper=\small,
    boxsep=2pt,
    left=3pt, right=3pt, top=2pt, bottom=2pt,
    title=2. College Physics
]
\renewcommand{\arraystretch}{1.15}
\begin{tabularx}{\linewidth}{@{}p{2.2cm} X@{}}
\toprule
\textbf{Version} & \textbf{Prompt} \\
\midrule
\colorbox{yellow!20}{\textbf{Simple}} &
You are an expert in the field of college physics. Please answer the following question based on your professional knowledge. \\
\colorbox{orange!20}{\textbf{Moderate}} &
You are a professional physics expert in college-level physics. Your expertise includes classical mechanics, thermodynamics, electromagnetism, waves, optics, and modern physics. You understand fundamental physical laws and core definitions. Please choose the answer based on rigorous physical principles. \\
\colorbox{violet!12}{\textbf{Rich}} &
You are a professional physics expert who masters the full range of college physics knowledge. Your expertise includes classical mechanics, fluid mechanics, thermodynamics, electromagnetism, mechanical waves, optics and basic modern physics. You have a solid understanding of fundamental physical laws, core definitions, physical models and calculation principles. You are able to identify confusing concepts and common mistakes in physics learning, and judge answers according to rigorous physical theories and objective rules. \\
\bottomrule
\end{tabularx}
\end{tcolorbox}

\vspace{1mm}

\begin{tcolorbox}[
    colback=gray!5!white,
    colframe=gray!50!black,
    coltitle=white,
    fonttitle=\bfseries\small,
    fontupper=\small,
    boxsep=2pt,
    left=3pt, right=3pt, top=2pt, bottom=2pt,
    title=3. High School Geography
]
\renewcommand{\arraystretch}{1.15}
\begin{tabularx}{\linewidth}{@{}p{2.2cm} X@{}}
\toprule
\textbf{Version} & \textbf{Prompt} \\
\midrule
\colorbox{yellow!20}{\textbf{Simple}} &
You are an expert in the field of high school geography. Please answer the following question based on your professional knowledge. \\
\colorbox{orange!20}{\textbf{Moderate}} &
You are a professional geography teacher. You are proficient in physical geography (terrain, climate, hydrology) and human geography (population, urban development, industry, agriculture). You understand how natural environments and human activities interact. Please select the answer according to geographical principles. \\
\colorbox{violet!12}{\textbf{Rich}} &
You are a professional geography teacher with solid theoretical knowledge. You are proficient in both physical geography and human geography. Your knowledge includes global terrain features, climate distribution, hydrological circulation, soil and vegetation, as well as population distribution, urban development, industry, agriculture, transportation and regional differences. You understand the mutual influence between human society and natural environment, and analyze problems from the perspective of geographical laws and regional characteristics. \\
\bottomrule
\end{tabularx}

\end{tcolorbox}

\caption{Comparison of role-playing prompts with differently detailed character information (1/3).}
\label{table:7}
\end{table*}

\begin{table*}[htbp]
\centering

\small

\begin{tcolorbox}[
    colback=gray!5!white,
    colframe=gray!50!black,
    coltitle=white,
    fonttitle=\bfseries\small,
    fontupper=\small,
    boxsep=2pt,
    left=3pt, right=3pt, top=2pt, bottom=2pt,
    title=4. Business Ethics
]
\renewcommand{\arraystretch}{1.15}
\begin{tabularx}{\linewidth}{@{}p{2.2cm} X@{}}
\toprule
\textbf{Version} & \textbf{Prompt} \\
\midrule
\colorbox{yellow!20}{\textbf{Simple}} &
You are an expert in the field of business ethics. Please answer the following question based on your professional knowledge. \\
\colorbox{orange!20}{\textbf{Moderate}} &
You are an expert in business ethics and corporate responsibility. You are familiar with ethical theories, compliance requirements, fair competition, labor rights, and corporate governance. You evaluate business practices from moral and professional standards. Please choose the answer that aligns with mainstream business ethics. \\
\colorbox{violet!12}{\textbf{Rich}} &
You are an expert focusing on business ethics and corporate social responsibility. You are familiar with basic ethical theories, commercial behavior norms and industry compliance requirements. Your work covers conflicts of interest, fair competition, labor rights, advertising credibility, environmental protection, data ethics and corporate governance. You can evaluate business behaviors from moral, legal and social perspectives, and select answers that conform to mainstream business ethics and professional standards. \\
\bottomrule
\end{tabularx}
\end{tcolorbox}

\vspace{1mm}

\begin{tcolorbox}[
    colback=gray!5!white,
    colframe=gray!50!black,
    coltitle=white,
    fonttitle=\bfseries\small,
    fontupper=\small,
    boxsep=2pt,
    left=3pt, right=3pt, top=2pt, bottom=2pt,
    title=5. Human Aging
]
\renewcommand{\arraystretch}{1.15}
\begin{tabularx}{\linewidth}{@{}p{2.2cm} X@{}}
\toprule
\textbf{Version} & \textbf{Prompt} \\
\midrule
\colorbox{yellow!20}{\textbf{Simple}} &
You are an expert in the field of human aging. Please answer the following question based on your professional knowledge. \\
\colorbox{orange!20}{\textbf{Moderate}} &
You are an expert in gerontology and human aging. You have solid knowledge of biological aging processes, psychological changes in later life, age-related diseases, and social aspects of aging. You understand theories of aging, longevity factors, and interventions for healthy aging. Please select the answer based on scientific evidence in gerontology. \\
\colorbox{violet!12}{\textbf{Rich}} &
You are a professional expert in gerontology and human aging. You have comprehensive knowledge of the biological, psychological, and social aspects of aging. Your expertise covers cellular senescence, age-related physiological changes, cognitive decline, neurodegenerative diseases, and the impact of lifestyle factors on longevity. You understand theories of aging, demographic trends in population aging, and interventions for healthy aging. Please analyze the question carefully based on scientific research and clinical evidence in gerontology. \\
\bottomrule
\end{tabularx}
\end{tcolorbox}

\vspace{1mm}

\begin{tcolorbox}[
    colback=gray!5!white,
    colframe=gray!50!black,
    coltitle=white,
    fonttitle=\bfseries\small,
    fontupper=\small,
    boxsep=2pt,
    left=3pt, right=3pt, top=2pt, bottom=2pt,
    title=6. Philosophy
]
\renewcommand{\arraystretch}{1.15}
\begin{tabularx}{\linewidth}{@{}p{2.2cm} X@{}}
\toprule
\textbf{Version} & \textbf{Prompt} \\
\midrule
\colorbox{yellow!20}{\textbf{Simple}} &
You are an expert in the field of philosophy. Please answer the following question based on your professional knowledge. \\
\colorbox{orange!20}{\textbf{Moderate}} &
You are a professional philosopher with knowledge across major philosophical traditions. Your expertise includes metaphysics, epistemology, ethics, logic, and political philosophy. You are familiar with key thinkers and philosophical arguments. Please reason through the question using rigorous philosophical analysis. \\
\colorbox{violet!12}{\textbf{Rich}} &
You are a professional philosopher with broad knowledge across major philosophical traditions. Your expertise includes metaphysics, epistemology, ethics, logic, philosophy of mind, philosophy of science, and political philosophy. You are familiar with key thinkers from ancient Greek philosophy to contemporary analytic and continental traditions. You understand philosophical argumentation, conceptual analysis, and can distinguish between different philosophical positions and their implications. Please reason through the question using rigorous philosophical analysis. \\
\bottomrule
\end{tabularx}
\end{tcolorbox}
\caption{Comparison of role-playing prompts with differently detailed character information (2/3).}
\label{table:8}
\end{table*}

\begin{table*}[t]
\centering

\small

\begin{tcolorbox}[
    colback=gray!5!white,
    colframe=gray!50!black,
    coltitle=white,
    fonttitle=\bfseries\small,
    fontupper=\small,
    boxsep=2pt,
    left=3pt, right=3pt, top=2pt, bottom=2pt,
    title=7. Professional Accounting
]
\renewcommand{\arraystretch}{1.15}
\begin{tabularx}{\linewidth}{@{}p{2.2cm} X@{}}
\toprule
\textbf{Version} & \textbf{Prompt} \\
\midrule
\colorbox{yellow!20}{\textbf{Simple}} &
You are an expert in the field of professional accounting. Please answer the following question based on your professional knowledge. \\
\colorbox{orange!20}{\textbf{Moderate}} &
You are a professional accountant with practical experience. You have knowledge of financial accounting standards, managerial accounting, auditing, tax regulations, and financial reporting. Your expertise covers GAAP, IFRS, cost accounting, and professional ethics. Please select the answer that follows professional accounting standards. \\
\colorbox{violet!12}{\textbf{Rich}} &
You are a professional certified accountant with extensive practical experience. You have comprehensive knowledge of financial accounting standards, managerial accounting, auditing procedures, tax regulations, and financial reporting requirements. Your expertise covers GAAP, IFRS, cost accounting, budgeting, internal controls, and professional ethics in accounting. You understand how to apply accounting principles to real business scenarios and can identify correct treatment of various financial transactions. Please select the answer that follows professional accounting standards. \\
\bottomrule
\end{tabularx}
\end{tcolorbox}
\caption{Comparison of role-playing prompts with differently detailed character information (3/3).}
\label{table:9}
\end{table*}

\renewcommand{\arraystretch}{0.85}
\setlength{\tabcolsep}{3pt}

\begin{table*}[t]
\centering
\footnotesize
\resizebox{\linewidth}{!}{
\begin{tabular}{lccccccc}
\toprule
\multirow{2}{*}{Subject} & \multirow{2}{*}{Setting} & \multicolumn{6}{c}{Models} \\
\cmidrule(lr){3-8}
 & & Mistral-7B & Llama3-8B & Qwen3-4B & Qwen3-8B & Qwen3-14B & Qwen3-32B \\
\midrule
\multirow{2}{*}{anatomy} & Base & 0.460 & 0.620 & 0.750 & 0.780 & 0.800 & 0.830 \\
 & Role-playing & 0.490 {\color{teal}(+3\%)} & 0.650 {\color{teal}(+3\%)} & 0.660 {\color{red}(-9\%)} & 0.720 {\color{red}(-6\%)} & 0.810 {\color{teal}(+1\%)} & 0.850 {\color{teal}(+2\%)} \\
\midrule
\multirow{2}{*}{astronomy} & Base & 0.610 & 0.680 & 0.860 & 0.910 & 0.940 & 0.950 \\
 & Role-playing & 0.640 {\color{teal}(+3\%)} & 0.650 {\color{red}(-3\%)} & 0.880 {\color{teal}(+2\%)} & 0.900 {\color{red}(-1\%)} & 0.920 {\color{red}(-2\%)} & 0.960 {\color{teal}(+1\%)} \\
\midrule
\multirow{2}{*}{business\_ethics} & Base & 0.590 & 0.600 & 0.820 & 0.790 & 0.840 & 0.860 \\
 & Role-playing & 0.600 {\color{teal}(+1\%)} & 0.630 {\color{teal}(+3\%)} & 0.770 {\color{red}(-5\%)} & 0.810 {\color{teal}(+2\%)} & 0.860 {\color{teal}(+2\%)} & 0.880 {\color{teal}(+2\%)} \\
\midrule
\multirow{2}{*}{clinical\_knowledge} & Base & 0.560 & 0.730 & 0.810 & 0.850 & 0.870 & 0.890 \\
 & Role-playing & 0.590 {\color{teal}(+3\%)} & 0.730 {\color{teal}(+0\%)} & 0.840 {\color{teal}(+3\%)} & 0.870 {\color{teal}(+2\%)} & 0.840 {\color{red}(-3\%)} & 0.910 {\color{teal}(+2\%)} \\
\midrule
\multirow{2}{*}{college\_chemistry} & Base & 0.340 & 0.370 & 0.590 & 0.630 & 0.680 & 0.720 \\
 & Role-playing & 0.360 {\color{teal}(+2\%)} & 0.400 {\color{teal}(+3\%)} & 0.570 {\color{red}(-2\%)} & 0.610 {\color{red}(-2\%)} & 0.650 {\color{red}(-3\%)} & 0.720 {\color{teal}(+0\%)} \\
\midrule
\multirow{2}{*}{college\_computer\_science} & Base & 0.440 & 0.460 & 0.740 & 0.670 & 0.790 & 0.820 \\
 & Role-playing & 0.430 {\color{red}(-1\%)} & 0.440 {\color{red}(-2\%)} & 0.760 {\color{teal}(+2\%)} & 0.730 {\color{teal}(+6\%)} & 0.800 {\color{teal}(+1\%)} & 0.840 {\color{teal}(+2\%)} \\
\midrule
\multirow{2}{*}{college\_mathematics} & Base & 0.280 & 0.360 & 0.670 & 0.710 & 0.720 & 0.760 \\
 & Role-playing & 0.270 {\color{red}(-1\%)} & 0.300 {\color{red}(-6\%)} & 0.710 {\color{teal}(+4\%)} & 0.650 {\color{red}(-6\%)} & 0.740 {\color{teal}(+2\%)} & 0.750 {\color{red}(-1\%)} \\
\midrule
\multirow{2}{*}{college\_medicine} & Base & 0.510 & 0.610 & 0.800 & 0.810 & 0.790 & 0.840 \\
 & Role-playing & 0.510 {\color{teal}(+0\%)} & 0.600 {\color{red}(-1\%)} & 0.810 {\color{teal}(+1\%)} & 0.800 {\color{red}(-1\%)} & 0.820 {\color{teal}(+3\%)} & 0.860 {\color{teal}(+2\%)} \\
\midrule
\multirow{2}{*}{college\_physics} & Base & 0.310 & 0.440 & 0.890 & 0.790 & 0.880 & 0.900 \\
 & Role-playing & 0.340 {\color{teal}(+3\%)} & 0.440 {\color{teal}(+0\%)} & 0.820 {\color{red}(-7\%)} & 0.840 {\color{teal}(+5\%)} & 0.910 {\color{teal}(+3\%)} & 0.920 {\color{teal}(+2\%)} \\
\midrule
\multirow{2}{*}{conceptual\_physics} & Base & 0.400 & 0.510 & 0.870 & 0.920 & 0.950 & 0.960 \\
 & Role-playing & 0.430 {\color{teal}(+3\%)} & 0.490 {\color{red}(-2\%)} & 0.880 {\color{teal}(+1\%)} & 0.930 {\color{teal}(+1\%)} & 0.950 {\color{teal}(+0\%)} & 0.970 {\color{teal}(+1\%)} \\
\midrule
\multirow{2}{*}{econometrics} & Base & 0.390 & 0.450 & 0.640 & 0.630 & 0.690 & 0.730 \\
 & Role-playing & 0.400 {\color{teal}(+1\%)} & 0.460 {\color{teal}(+1\%)} & 0.660 {\color{teal}(+2\%)} & 0.610 {\color{red}(-2\%)} & 0.750 {\color{teal}(+6\%)} & 0.750 {\color{teal}(+2\%)} \\
\midrule
\multirow{2}{*}{electrical\_engineering} & Base & 0.510 & 0.520 & 0.770 & 0.820 & 0.860 & 0.880 \\
 & Role-playing & 0.480 {\color{red}(-3\%)} & 0.530 {\color{teal}(+1\%)} & 0.820 {\color{teal}(+5\%)} & 0.840 {\color{teal}(+2\%)} & 0.890 {\color{teal}(+3\%)} & 0.900 {\color{teal}(+2\%)} \\
\midrule
\multirow{2}{*}{formal\_logic} & Base & 0.430 & 0.420 & 0.850 & 0.780 & 0.910 & 0.920 \\
 & Role-playing & 0.370 {\color{red}(-6\%)} & 0.390 {\color{red}(-3\%)} & 0.890 {\color{teal}(+4\%)} & 0.820 {\color{teal}(+4\%)} & 0.880 {\color{red}(-3\%)} & 0.930 {\color{teal}(+1\%)} \\
\midrule
\multirow{2}{*}{global\_facts} & Base & 0.230 & 0.430 & 0.430 & 0.510 & 0.510 & 0.560 \\
 & Role-playing & 0.260 {\color{teal}(+3\%)} & 0.390 {\color{red}(-4\%)} & 0.430 {\color{teal}(+0\%)} & 0.500 {\color{red}(-1\%)} & 0.510 {\color{teal}(+0\%)} & 0.550 {\color{red}(-1\%)} \\
\midrule
\multirow{2}{*}{high\_school\_chemistry} & Base & 0.320 & 0.430 & 0.820 & 0.800 & 0.860 & 0.880 \\
 & Role-playing & 0.300 {\color{red}(-2\%)} & 0.370 {\color{red}(-6\%)} & 0.810 {\color{red}(-1\%)} & 0.770 {\color{red}(-3\%)} & 0.890 {\color{teal}(+3\%)} & 0.880 {\color{teal}(+0\%)} \\
\bottomrule
\end{tabular}
}
\caption{Performance comparison of base vs. role-playing modes across different models and subjects on MMLU-Redux (Part 1/2).}
\label{table:10}
\end{table*}

\begin{table*}[t]
\centering
\footnotesize
\resizebox{\linewidth}{!}{
\begin{tabular}{lccccccc}
\toprule
\multirow{2}{*}{Subject} & \multirow{2}{*}{Setting} & \multicolumn{6}{c}{Models} \\
\cmidrule(lr){3-8}
 & & Mistral-7B & Llama3-8B & Qwen3-4B & Qwen3-8B & Qwen3-14B & Qwen3-32B \\
\midrule
\multirow{2}{*}{high\_school\_geography} & Base & 0.600 & 0.770 & 0.880 & 0.870 & 0.870 & 0.900 \\
 & Role-playing & 0.610 {\color{teal}(+1\%)} & 0.720 {\color{red}(-5\%)} & 0.820 {\color{red}(-6\%)} & 0.870 {\color{teal}(+0\%)} & 0.900 {\color{teal}(+3\%)} & 0.910 {\color{teal}(+1\%)} \\
\midrule
\multirow{2}{*}{high\_school\_macroeconomics} & Base & 0.380 & 0.460 & 0.770 & 0.810 & 0.890 & 0.900 \\
 & Role-playing & 0.390 {\color{teal}(+1\%)} & 0.450 {\color{red}(-1\%)} & 0.770 {\color{teal}(+0\%)} & 0.850 {\color{teal}(+4\%)} & 0.880 {\color{red}(-1\%)} & 0.910 {\color{teal}(+1\%)} \\
\midrule
\multirow{2}{*}{high\_school\_mathematics} & Base & 0.320 & 0.380 & 0.750 & 0.740 & 0.740 & 0.780 \\
 & Role-playing & 0.300 {\color{red}(-2\%)} & 0.360 {\color{red}(-2\%)} & 0.780 {\color{teal}(+3\%)} & 0.690 {\color{red}(-5\%)} & 0.750 {\color{teal}(+1\%)} & 0.790 {\color{teal}(+1\%)} \\
\midrule
\multirow{2}{*}{high\_school\_physics} & Base & 0.230 & 0.420 & 0.810 & 0.830 & 0.860 & 0.890 \\
 & Role-playing & 0.270 {\color{teal}(+4\%)} & 0.360 {\color{red}(-6\%)} & 0.810 {\color{teal}(+0\%)} & 0.830 {\color{teal}(+0\%)} & 0.890 {\color{teal}(+3\%)} & 0.900 {\color{teal}(+1\%)} \\
\midrule
\multirow{2}{*}{high\_school\_statistics} & Base & 0.400 & 0.450 & 0.860 & 0.800 & 0.870 & 0.890 \\
 & Role-playing & 0.400 {\color{teal}(+0\%)} & 0.390 {\color{red}(-6\%)} & 0.850 {\color{red}(-1\%)} & 0.830 {\color{teal}(+3\%)} & 0.900 {\color{teal}(+3\%)} & 0.900 {\color{teal}(+1\%)} \\
\midrule
\multirow{2}{*}{high\_school\_us\_history} & Base & 0.710 & 0.740 & 0.900 & 0.920 & 0.890 & 0.930 \\
 & Role-playing & 0.730 {\color{teal}(+2\%)} & 0.750 {\color{teal}(+1\%)} & 0.880 {\color{red}(-2\%)} & 0.940 {\color{teal}(+2\%)} & 0.910 {\color{teal}(+2\%)} & 0.950 {\color{teal}(+2\%)} \\
\midrule
\multirow{2}{*}{human\_aging} & Base & 0.550 & 0.590 & 0.750 & 0.730 & 0.740 & 0.780 \\
 & Role-playing & 0.550 {\color{teal}(+0\%)} & 0.600 {\color{teal}(+1\%)} & 0.710 {\color{red}(-4\%)} & 0.770 {\color{teal}(+4\%)} & 0.730 {\color{red}(-1\%)} & 0.780 {\color{teal}(+0\%)} \\
\midrule
\multirow{2}{*}{logical\_fallacies} & Base & 0.670 & 0.720 & 0.880 & 0.900 & 0.860 & 0.910 \\
 & Role-playing & 0.680 {\color{teal}(+1\%)} & 0.730 {\color{teal}(+1\%)} & 0.910 {\color{teal}(+3\%)} & 0.910 {\color{teal}(+1\%)} & 0.860 {\color{teal}(+0\%)} & 0.930 {\color{teal}(+2\%)} \\
\midrule
\multirow{2}{*}{machine\_learning} & Base & 0.500 & 0.490 & 0.750 & 0.770 & 0.850 & 0.870 \\
 & Role-playing & 0.440 {\color{red}(-6\%)} & 0.480 {\color{red}(-1\%)} & 0.760 {\color{teal}(+1\%)} & 0.760 {\color{red}(-1\%)} & 0.870 {\color{teal}(+2\%)} & 0.890 {\color{teal}(+2\%)} \\
\midrule
\multirow{2}{*}{miscellaneous} & Base & 0.680 & 0.810 & 0.890 & 0.910 & 0.940 & 0.950 \\
 & Role-playing & 0.710 {\color{teal}(+3\%)} & 0.800 {\color{red}(-1\%)} & 0.890 {\color{teal}(+0\%)} & 0.950 {\color{teal}(+4\%)} & 0.940 {\color{teal}(+0\%)} & 0.970 {\color{teal}(+2\%)} \\
\midrule
\multirow{2}{*}{philosophy} & Base & 0.530 & 0.660 & 0.730 & 0.700 & 0.770 & 0.800 \\
 & Role-playing & 0.600 {\color{teal}(+7\%)} & 0.650 {\color{red}(-1\%)} & 0.690 {\color{red}(-4\%)} & 0.750 {\color{teal}(+5\%)} & 0.800 {\color{teal}(+3\%)} & 0.830 {\color{teal}(+3\%)} \\
\midrule
\multirow{2}{*}{professional\_accounting} & Base & 0.390 & 0.430 & 0.680 & 0.770 & 0.730 & 0.780 \\
 & Role-playing & 0.420 {\color{teal}(+3\%)} & 0.420 {\color{red}(-1\%)} & 0.640 {\color{red}(-4\%)} & 0.680 {\color{red}(-9\%)} & 0.760 {\color{teal}(+3\%)} & 0.780 {\color{teal}(+0\%)} \\
\midrule
\multirow{2}{*}{professional\_law} & Base & 0.370 & 0.450 & 0.490 & 0.520 & 0.550 & 0.600 \\
 & Role-playing & 0.420 {\color{teal}(+5\%)} & 0.400 {\color{red}(-5\%)} & 0.520 {\color{teal}(+3\%)} & 0.480 {\color{red}(-4\%)} & 0.610 {\color{teal}(+6\%)} & 0.590 {\color{red}(-1\%)} \\
\midrule
\multirow{2}{*}{public\_relations} & Base & 0.560 & 0.670 & 0.730 & 0.730 & 0.760 & 0.790 \\
 & Role-playing & 0.540 {\color{red}(-2\%)} & 0.650 {\color{red}(-2\%)} & 0.720 {\color{red}(-1\%)} & 0.760 {\color{teal}(+3\%)} & 0.760 {\color{teal}(+0\%)} & 0.780 {\color{red}(-1\%)} \\
\midrule
\multirow{2}{*}{virology} & Base & 0.460 & 0.510 & 0.500 & 0.510 & 0.520 & 0.560 \\
 & Role-playing & 0.470 {\color{teal}(+1\%)} & 0.490 {\color{red}(-2\%)} & 0.550 {\color{teal}(+5\%)} & 0.550 {\color{teal}(+4\%)} & 0.530 {\color{teal}(+1\%)} & 0.570 {\color{teal}(+1\%)} \\
\bottomrule
\end{tabular}
}
\caption{Performance comparison of base vs. role-playing modes across different models and subjects on MMLU-Redux (Part 2/2).}
\label{table:11}
\end{table*}

\renewcommand{\arraystretch}{0.85}
\setlength{\tabcolsep}{3pt}
\begin{table*}[t]
\centering
\footnotesize
\resizebox{\linewidth}{!}{
\begin{tabular}{lccccccc}
\toprule
\multirow{2}{*}{Subject} & \multirow{2}{*}{Setting} & \multicolumn{6}{c}{Models} \\
\cmidrule(lr){3-8}
 & & Mistral-7B & Llama3-8B & Qwen3-4B & Qwen3-8B & Qwen3-14B & Qwen3-32B \\
\midrule
\multirow{2}{*}{abstract\_algebra} & Base & 0.270 & 0.350 & 0.780 & 0.790 & 0.840 & 0.873 \\
 & Role-playing & 0.280 {\color{teal}(+1.0\%)} & 0.370 {\color{teal}(+2.0\%)} & 0.770 {\color{red}(-1.0\%)} & 0.800 {\color{teal}(+1.0\%)} & 0.830 {\color{red}(-1.0\%)} & 0.886 {\color{teal}(+1.3\%)} \\
\midrule
\multirow{2}{*}{anatomy} & Base & 0.533 & 0.630 & 0.696 & 0.748 & 0.837 & 0.866 \\
 & Role-playing & 0.570 {\color{teal}(+3.7\%)} & 0.659 {\color{teal}(+3.0\%)} & 0.667 {\color{red}(-3.0\%)} & 0.726 {\color{red}(-2.2\%)} & 0.800 {\color{red}(-3.7\%)} & 0.879 {\color{teal}(+1.3\%)} \\
\midrule
\multirow{2}{*}{astronomy} & Base & 0.651 & 0.658 & 0.842 & 0.928 & 0.901 & 0.934 \\
 & Role-playing & 0.632 {\color{red}(-2.0\%)} & 0.697 {\color{teal}(+3.9\%)} & 0.822 {\color{red}(-2.0\%)} & 0.901 {\color{red}(-2.6\%)} & 0.947 {\color{teal}(+4.6\%)} & 0.941 {\color{teal}(+0.7\%)} \\
\midrule
\multirow{2}{*}{business\_ethics} & Base & 0.620 & 0.660 & 0.790 & 0.760 & 0.810 & 0.842 \\
 & Role-playing & 0.620 {\color{teal}(+0.0\%)} & 0.600 {\color{red}(-6.0\%)} & 0.760 {\color{red}(-3.0\%)} & 0.800 {\color{teal}(+4.0\%)} & 0.800 {\color{red}(-1.0\%)} & 0.842 {\color{teal}(+0.0\%)} \\
\midrule
\multirow{2}{*}{clinical\_knowledge} & Base & 0.630 & 0.683 & 0.792 & 0.815 & 0.860 & 0.885 \\
 & Role-playing & 0.630 {\color{teal}(+0.0\%)} & 0.683 {\color{teal}(+0.0\%)} & 0.815 {\color{teal}(+2.3\%)} & 0.823 {\color{teal}(+0.8\%)} & 0.860 {\color{teal}(+0.0\%)} & 0.904 {\color{teal}(+1.9\%)} \\
\midrule
\multirow{2}{*}{college\_biology} & Base & 0.681 & 0.708 & 0.854 & 0.896 & 0.944 & 0.947 \\
 & Role-playing & 0.674 {\color{red}(-0.7\%)} & 0.701 {\color{red}(-0.7\%)} & 0.910 {\color{teal}(+5.6\%)} & 0.910 {\color{teal}(+1.4\%)} & 0.951 {\color{teal}(+0.7\%)} & 0.958 {\color{teal}(+1.1\%)} \\
\midrule
\multirow{2}{*}{college\_chemistry} & Base & 0.460 & 0.350 & 0.620 & 0.620 & 0.700 & 0.734 \\
 & Role-playing & 0.490 {\color{teal}(+3.0\%)} & 0.390 {\color{teal}(+4.0\%)} & 0.650 {\color{teal}(+3.0\%)} & 0.680 {\color{teal}(+6.0\%)} & 0.660 {\color{red}(-4.0\%)} & 0.726 {\color{red}(-0.8\%)} \\
\midrule
\multirow{2}{*}{college\_computer\_science} & Base & 0.490 & 0.450 & 0.740 & 0.690 & 0.830 & 0.853 \\
 & Role-playing & 0.450 {\color{red}(-4.0\%)} & 0.420 {\color{red}(-3.0\%)} & 0.720 {\color{red}(-2.0\%)} & 0.710 {\color{teal}(+2.0\%)} & 0.800 {\color{red}(-3.0\%)} & 0.871 {\color{teal}(+1.8\%)} \\
\midrule
\multirow{2}{*}{college\_mathematics} & Base & 0.250 & 0.230 & 0.680 & 0.710 & 0.740 & 0.776 \\
 & Role-playing & 0.300 {\color{teal}(+5.0\%)} & 0.290 {\color{teal}(+6.0\%)} & 0.690 {\color{teal}(+1.0\%)} & 0.640 {\color{red}(-7.0\%)} & 0.740 {\color{teal}(+0.0\%)} & 0.789 {\color{teal}(+1.3\%)} \\
\midrule
\multirow{2}{*}{college\_medicine} & Base & 0.555 & 0.601 & 0.786 & 0.775 & 0.815 & 0.848 \\
 & Role-playing & 0.561 {\color{teal}(+0.6\%)} & 0.566 {\color{red}(-3.5\%)} & 0.786 {\color{teal}(+0.0\%)} & 0.798 {\color{teal}(+2.3\%)} & 0.815 {\color{teal}(+0.0\%)} & 0.863 {\color{teal}(+1.5\%)} \\
\midrule
\multirow{2}{*}{college\_physics} & Base & 0.402 & 0.382 & 0.843 & 0.873 & 0.882 & 0.904 \\
 & Role-playing & 0.382 {\color{red}(-2.0\%)} & 0.392 {\color{teal}(+1.0\%)} & 0.863 {\color{teal}(+2.0\%)} & 0.843 {\color{red}(-2.9\%)} & 0.922 {\color{teal}(+3.9\%)} & 0.917 {\color{teal}(+1.3\%)} \\
\midrule
\multirow{2}{*}{computer\_security} & Base & 0.660 & 0.730 & 0.800 & 0.760 & 0.830 & 0.857 \\
 & Role-playing & 0.670 {\color{teal}(+1.0\%)} & 0.730 {\color{teal}(+0.0\%)} & 0.790 {\color{red}(-1.0\%)} & 0.800 {\color{teal}(+4.0\%)} & 0.860 {\color{teal}(+3.0\%)} & 0.878 {\color{teal}(+2.1\%)} \\
\midrule
\multirow{2}{*}{conceptual\_physics} & Base & 0.489 & 0.579 & 0.864 & 0.919 & 0.940 & 0.958 \\
 & Role-playing & 0.494 {\color{teal}(+0.4\%)} & 0.579 {\color{teal}(+0.0\%)} & 0.868 {\color{teal}(+0.4\%)} & 0.911 {\color{red}(-0.9\%)} & 0.949 {\color{teal}(+0.9\%)} & 0.969 {\color{teal}(+1.1\%)} \\
\midrule
\multirow{2}{*}{econometrics} & Base & 0.377 & 0.456 & 0.632 & 0.702 & 0.711 & 0.763 \\
 & Role-playing & 0.439 {\color{teal}(+6.1\%)} & 0.474 {\color{teal}(+1.8\%)} & 0.684 {\color{teal}(+5.3\%)} & 0.632 {\color{red}(-7.0\%)} & 0.746 {\color{teal}(+3.5\%)} & 0.784 {\color{teal}(+2.1\%)} \\
\midrule
\multirow{2}{*}{electrical\_engineering} & Base & 0.469 & 0.538 & 0.793 & 0.828 & 0.855 & 0.884 \\
 & Role-playing & 0.503 {\color{teal}(+3.4\%)} & 0.586 {\color{teal}(+4.8\%)} & 0.786 {\color{red}(-0.7\%)} & 0.793 {\color{red}(-3.4\%)} & 0.869 {\color{teal}(+1.4\%)} & 0.903 {\color{teal}(+1.9\%)} \\
\midrule
\multirow{2}{*}{elementary\_mathematics} & Base & 0.394 & 0.410 & 0.950 & 0.963 & 0.979 & 0.983 \\
 & Role-playing & 0.378 {\color{red}(-1.6\%)} & 0.426 {\color{teal}(+1.6\%)} & 0.955 {\color{teal}(+0.5\%)} & 0.966 {\color{teal}(+0.3\%)} & 0.976 {\color{red}(-0.3\%)} & 0.983 {\color{teal}(+0.0\%)} \\
\midrule
\multirow{2}{*}{formal\_logic} & Base & 0.476 & 0.429 & 0.746 & 0.794 & 0.857 & 0.876 \\
 & Role-playing & 0.452 {\color{red}(-2.4\%)} & 0.429 {\color{teal}(+0.0\%)} & 0.770 {\color{teal}(+2.4\%)} & 0.722 {\color{red}(-7.1\%)} & 0.881 {\color{teal}(+2.4\%)} & 0.892 {\color{teal}(+1.6\%)} \\
\midrule
\multirow{2}{*}{global\_facts} & Base & 0.370 & 0.370 & 0.450 & 0.500 & 0.560 & 0.604 \\
 & Role-playing & 0.370 {\color{teal}(+0.0\%)} & 0.360 {\color{red}(-1.0\%)} & 0.520 {\color{teal}(+7.0\%)} & 0.540 {\color{teal}(+4.0\%)} & 0.540 {\color{red}(-2.0\%)} & 0.604 {\color{teal}(+0.0\%)} \\
\midrule
\multirow{2}{*}{high\_school\_biology} & Base & 0.729 & 0.758 & 0.890 & 0.906 & 0.926 & 0.938 \\
 & Role-playing & 0.729 {\color{teal}(+0.0\%)} & 0.771 {\color{teal}(+1.3\%)} & 0.881 {\color{red}(-1.0\%)} & 0.900 {\color{red}(-0.6\%)} & 0.929 {\color{teal}(+0.3\%)} & 0.952 {\color{teal}(+1.4\%)} \\
\midrule
\multirow{2}{*}{high\_school\_chemistry} & Base & 0.463 & 0.448 & 0.803 & 0.818 & 0.847 & 0.873 \\
 & Role-playing & 0.443 {\color{red}(-2.0\%)} & 0.443 {\color{red}(-0.5\%)} & 0.773 {\color{red}(-3.0\%)} & 0.813 {\color{red}(-0.5\%)} & 0.857 {\color{teal}(+1.0\%)} & 0.885 {\color{teal}(+1.2\%)} \\
\midrule
\multirow{2}{*}{high\_school\_computer\_science} & Base & 0.590 & 0.660 & 0.900 & 0.920 & 0.940 & 0.953 \\
 & Role-playing & 0.560 {\color{red}(-3.0\%)} & 0.620 {\color{red}(-4.0\%)} & 0.910 {\color{teal}(+1.0\%)} & 0.930 {\color{teal}(+1.0\%)} & 0.960 {\color{teal}(+2.0\%)} & 0.964 {\color{teal}(+1.1\%)} \\
\midrule
\multirow{2}{*}{high\_school\_european\_history} & Base & 0.733 & 0.758 & 0.794 & 0.848 & 0.861 & 0.884 \\
 & Role-playing & 0.739 {\color{teal}(+0.6\%)} & 0.758 {\color{teal}(+0.0\%)} & 0.806 {\color{teal}(+1.2\%)} & 0.830 {\color{red}(-1.8\%)} & 0.867 {\color{teal}(+0.6\%)} & 0.897 {\color{teal}(+1.3\%)} \\
\midrule
\multirow{2}{*}{high\_school\_geography} & Base & 0.742 & 0.758 & 0.843 & 0.879 & 0.909 & 0.923 \\
 & Role-playing & 0.722 {\color{red}(-2.0\%)} & 0.753 {\color{red}(-0.5\%)} & 0.848 {\color{teal}(+0.5\%)} & 0.884 {\color{teal}(+0.5\%)} & 0.909 {\color{teal}(+0.0\%)} & 0.936 {\color{teal}(+1.3\%)} \\
\midrule
\multirow{2}{*}{high\_school\_government\_and\_politics} & Base & 0.824 & 0.772 & 0.891 & 0.933 & 0.959 & 0.957 \\
 & Role-playing & 0.813 {\color{red}(-1.0\%)} & 0.813 {\color{teal}(+4.1\%)} & 0.902 {\color{teal}(+1.0\%)} & 0.938 {\color{teal}(+0.5\%)} & 0.959 {\color{teal}(+0.0\%)} & 0.968 {\color{teal}(+1.1\%)} \\
\midrule
\multirow{2}{*}{high\_school\_macroeconomics} & Base & 0.533 & 0.615 & 0.854 & 0.874 & 0.910 & 0.924 \\
 & Role-playing & 0.551 {\color{teal}(+1.8\%)} & 0.656 {\color{teal}(+4.1\%)} & 0.846 {\color{red}(-0.8\%)} & 0.864 {\color{red}(-1.0\%)} & 0.905 {\color{red}(-0.5\%)} & 0.935 {\color{teal}(+1.1\%)} \\
\midrule
\multirow{2}{*}{high\_school\_mathematics} & Base & 0.307 & 0.344 & 0.796 & 0.796 & 0.789 & 0.823 \\
 & Role-playing & 0.300 {\color{red}(-0.7\%)} & 0.341 {\color{red}(-0.4\%)} & 0.811 {\color{teal}(+1.5\%)} & 0.759 {\color{red}(-3.7\%)} & 0.785 {\color{red}(-0.4\%)} & 0.836 {\color{teal}(+1.3\%)} \\
\midrule
\multirow{2}{*}{high\_school\_microeconomics} & Base & 0.592 & 0.655 & 0.924 & 0.933 & 0.958 & 0.963 \\
 & Role-playing & 0.597 {\color{teal}(+0.4\%)} & 0.710 {\color{teal}(+5.5\%)} & 0.937 {\color{teal}(+1.3\%)} & 0.937 {\color{teal}(+0.4\%)} & 0.962 {\color{teal}(+0.4\%)} & 0.974 {\color{teal}(+1.1\%)} \\
\midrule
\multirow{2}{*}{high\_school\_physics} & Base & 0.311 & 0.377 & 0.801 & 0.781 & 0.868 & 0.886 \\
 & Role-playing & 0.325 {\color{teal}(+1.3\%)} & 0.318 {\color{red}(-6.0\%)} & 0.795 {\color{red}(-0.7\%)} & 0.781 {\color{teal}(+0.0\%)} & 0.881 {\color{teal}(+1.3\%)} & 0.902 {\color{teal}(+1.6\%)} \\
\midrule
\multirow{2}{*}{high\_school\_psychology} & Base & 0.780 & 0.800 & 0.910 & 0.934 & 0.934 & 0.944 \\
 & Role-playing & 0.806 {\color{teal}(+2.6\%)} & 0.789 {\color{red}(-1.1\%)} & 0.906 {\color{red}(-0.4\%)} & 0.930 {\color{red}(-0.4\%)} & 0.943 {\color{teal}(+0.9\%)} & 0.956 {\color{teal}(+1.2\%)} \\
\bottomrule
\end{tabular}
}
\caption{Performance comparison of base vs. role-playing modes across different models and subjects on MMLU (Part 1/3).}
\label{table:12}
\end{table*}

\renewcommand{\arraystretch}{0.85}
\setlength{\tabcolsep}{3pt}
\begin{table*}[t]
\centering
\footnotesize
\resizebox{\linewidth}{!}{
\begin{tabular}{lccccccc}
\toprule
\multirow{2}{*}{Subject} & \multirow{2}{*}{Setting} & \multicolumn{6}{c}{Models} \\
\cmidrule(lr){3-8}
 & & Mistral-7B & Llama3-8B & Qwen3-4B & Qwen3-8B & Qwen3-14B & Qwen3-32B \\
\midrule
\multirow{2}{*}{high\_school\_statistics} & Base & 0.431 & 0.421 & 0.810 & 0.843 & 0.884 & 0.903 \\
 & Role-playing & 0.454 {\color{teal}(+2.3\%)} & 0.417 {\color{red}(-0.5\%)} & 0.838 {\color{teal}(+2.8\%)} & 0.843 {\color{teal}(+0.0\%)} & 0.898 {\color{teal}(+1.4\%)} & 0.914 {\color{teal}(+1.1\%)} \\
\midrule
\multirow{2}{*}{high\_school\_us\_history} & Base & 0.740 & 0.750 & 0.873 & 0.892 & 0.917 & 0.934 \\
 & Role-playing & 0.775 {\color{teal}(+3.4\%)} & 0.750 {\color{teal}(+0.0\%)} & 0.843 {\color{red}(-2.9\%)} & 0.892 {\color{teal}(+0.0\%)} & 0.907 {\color{red}(-1.0\%)} & 0.946 {\color{teal}(+1.2\%)} \\
\midrule
\multirow{2}{*}{high\_school\_world\_history} & Base & 0.772 & 0.806 & 0.823 & 0.861 & 0.886 & 0.904 \\
 & Role-playing & 0.781 {\color{teal}(+0.8\%)} & 0.793 {\color{red}(-1.3\%)} & 0.840 {\color{teal}(+1.7\%)} & 0.865 {\color{teal}(+0.4\%)} & 0.895 {\color{teal}(+0.8\%)} & 0.916 {\color{teal}(+1.2\%)} \\
\midrule
\multirow{2}{*}{human\_aging} & Base & 0.709 & 0.664 & 0.735 & 0.744 & 0.771 & 0.803 \\
 & Role-playing & 0.677 {\color{red}(-3.1\%)} & 0.691 {\color{teal}(+2.7\%)} & 0.700 {\color{red}(-3.6\%)} & 0.762 {\color{teal}(+1.8\%)} & 0.807 {\color{teal}(+3.6\%)} & 0.794 {\color{red}(-0.9\%)} \\
\midrule
\multirow{2}{*}{human\_sexuality} & Base & 0.695 & 0.756 & 0.824 & 0.855 & 0.863 & 0.883 \\
 & Role-playing & 0.687 {\color{red}(-0.8\%)} & 0.718 {\color{red}(-3.8\%)} & 0.802 {\color{red}(-2.3\%)} & 0.870 {\color{teal}(+1.5\%)} & 0.870 {\color{teal}(+0.8\%)} & 0.895 {\color{teal}(+1.2\%)} \\
\midrule
\multirow{2}{*}{international\_law} & Base & 0.702 & 0.727 & 0.760 & 0.785 & 0.860 & 0.884 \\
 & Role-playing & 0.711 {\color{teal}(+0.8\%)} & 0.727 {\color{teal}(+0.0\%)} & 0.777 {\color{teal}(+1.7\%)} & 0.826 {\color{teal}(+4.1\%)} & 0.868 {\color{teal}(+0.8\%)} & 0.896 {\color{teal}(+1.2\%)} \\
\midrule
\multirow{2}{*}{jurisprudence} & Base & 0.713 & 0.657 & 0.806 & 0.843 & 0.852 & 0.874 \\
 & Role-playing & 0.704 {\color{red}(-0.9\%)} & 0.722 {\color{teal}(+6.5\%)} & 0.806 {\color{teal}(+0.0\%)} & 0.833 {\color{red}(-0.9\%)} & 0.852 {\color{teal}(+0.0\%)} & 0.874 {\color{teal}(+0.0\%)} \\
\midrule
\multirow{2}{*}{logical\_fallacies} & Base & 0.681 & 0.712 & 0.822 & 0.847 & 0.847 & 0.873 \\
 & Role-playing & 0.681 {\color{teal}(+0.0\%)} & 0.742 {\color{teal}(+3.1\%)} & 0.834 {\color{teal}(+1.2\%)} & 0.859 {\color{teal}(+1.2\%)} & 0.834 {\color{red}(-1.2\%)} & 0.886 {\color{teal}(+1.3\%)} \\
\midrule
\multirow{2}{*}{machine\_learning} & Base & 0.509 & 0.402 & 0.768 & 0.741 & 0.804 & 0.843 \\
 & Role-playing & 0.500 {\color{red}(-0.9\%)} & 0.420 {\color{teal}(+1.8\%)} & 0.723 {\color{red}(-4.5\%)} & 0.741 {\color{teal}(+0.0\%)} & 0.866 {\color{teal}(+6.2\%)} & 0.856 {\color{teal}(+1.3\%)} \\
\midrule
\multirow{2}{*}{management} & Base & 0.748 & 0.748 & 0.835 & 0.883 & 0.913 & 0.923 \\
 & Role-playing & 0.738 {\color{red}(-1.0\%)} & 0.777 {\color{teal}(+2.9\%)} & 0.854 {\color{teal}(+1.9\%)} & 0.883 {\color{teal}(+0.0\%)} & 0.913 {\color{teal}(+0.0\%)} & 0.934 {\color{teal}(+1.1\%)} \\
\midrule
\multirow{2}{*}{marketing} & Base & 0.821 & 0.868 & 0.897 & 0.902 & 0.919 & 0.934 \\
 & Role-playing & 0.808 {\color{red}(-1.3\%)} & 0.876 {\color{teal}(+0.9\%)} & 0.893 {\color{red}(-0.4\%)} & 0.919 {\color{teal}(+1.7\%)} & 0.919 {\color{teal}(+0.0\%)} & 0.945 {\color{teal}(+1.1\%)} \\
\midrule
\multirow{2}{*}{medical\_genetics} & Base & 0.630 & 0.740 & 0.870 & 0.910 & 0.980 & 0.982 \\
 & Role-playing & 0.610 {\color{red}(-2.0\%)} & 0.750 {\color{teal}(+1.0\%)} & 0.860 {\color{red}(-1.0\%)} & 0.850 {\color{red}(-6.0\%)} & 0.970 {\color{red}(-1.0\%)} & 0.982 {\color{teal}(+0.0\%)} \\
\midrule
\multirow{2}{*}{miscellaneous} & Base & 0.794 & 0.803 & 0.876 & 0.904 & 0.934 & 0.944 \\
 & Role-playing & 0.794 {\color{teal}(+0.0\%)} & 0.794 {\color{red}(-0.9\%)} & 0.872 {\color{red}(-0.4\%)} & 0.908 {\color{teal}(+0.4\%)} & 0.935 {\color{teal}(+0.1\%)} & 0.956 {\color{teal}(+1.2\%)} \\
\midrule
\multirow{2}{*}{moral\_disputes} & Base & 0.645 & 0.662 & 0.697 & 0.694 & 0.766 & 0.804 \\
 & Role-playing & 0.633 {\color{red}(-1.2\%)} & 0.688 {\color{teal}(+2.6\%)} & 0.685 {\color{red}(-1.2\%)} & 0.717 {\color{teal}(+2.3\%)} & 0.760 {\color{red}(-0.6\%)} & 0.793 {\color{red}(-1.1\%)} \\
\midrule
\multirow{2}{*}{moral\_scenarios} & Base & 0.383 & 0.347 & 0.641 & 0.666 & 0.699 & 0.734 \\
 & Role-playing & 0.417 {\color{teal}(+3.4\%)} & 0.293 {\color{red}(-5.5\%)} & 0.634 {\color{red}(-0.8\%)} & 0.666 {\color{teal}(+0.0\%)} & 0.705 {\color{teal}(+0.6\%)} & 0.746 {\color{teal}(+1.2\%)} \\
\midrule
\multirow{2}{*}{nutrition} & Base & 0.650 & 0.725 & 0.778 & 0.840 & 0.876 & 0.893 \\
 & Role-playing & 0.621 {\color{red}(-2.9\%)} & 0.755 {\color{teal}(+2.9\%)} & 0.784 {\color{teal}(+0.7\%)} & 0.827 {\color{red}(-1.3\%)} & 0.866 {\color{red}(-1.0\%)} & 0.905 {\color{teal}(+1.2\%)} \\
\midrule
\multirow{2}{*}{philosophy} & Base & 0.637 & 0.617 & 0.701 & 0.714 & 0.775 & 0.803 \\
 & Role-playing & 0.666 {\color{teal}(+2.9\%)} & 0.672 {\color{teal}(+5.5\%)} & 0.707 {\color{teal}(+0.6\%)} & 0.736 {\color{teal}(+2.3\%)} & 0.752 {\color{red}(-2.3\%)} & 0.814 {\color{teal}(+1.1\%)} \\
\midrule
\multirow{2}{*}{prehistory} & Base & 0.685 & 0.701 & 0.772 & 0.843 & 0.861 & 0.883 \\
 & Role-playing & 0.670 {\color{red}(-1.5\%)} & 0.719 {\color{teal}(+1.9\%)} & 0.762 {\color{red}(-0.9\%)} & 0.830 {\color{red}(-1.2\%)} & 0.877 {\color{teal}(+1.5\%)} & 0.896 {\color{teal}(+1.3\%)} \\
\midrule
\multirow{2}{*}{professional\_accounting} & Base & 0.447 & 0.511 & 0.709 & 0.780 & 0.844 & 0.863 \\
 & Role-playing & 0.447 {\color{teal}(+0.0\%)} & 0.493 {\color{red}(-1.8\%)} & 0.716 {\color{teal}(+0.7\%)} & 0.812 {\color{teal}(+3.2\%)} & 0.812 {\color{red}(-3.2\%)} & 0.875 {\color{teal}(+1.2\%)} \\
\midrule
\multirow{2}{*}{professional\_law} & Base & 0.423 & 0.453 & 0.465 & 0.492 & 0.581 & 0.624 \\
 & Role-playing & 0.427 {\color{teal}(+0.4\%)} & 0.443 {\color{red}(-1.0\%)} & 0.449 {\color{red}(-1.6\%)} & 0.483 {\color{red}(-0.9\%)} & 0.571 {\color{red}(-1.0\%)} & 0.613 {\color{red}(-1.1\%)} \\
\midrule
\multirow{2}{*}{professional\_medicine} & Base & 0.614 & 0.713 & 0.824 & 0.871 & 0.901 & 0.923 \\
 & Role-playing & 0.632 {\color{teal}(+1.8\%)} & 0.695 {\color{red}(-1.8\%)} & 0.805 {\color{red}(-1.8\%)} & 0.849 {\color{red}(-2.2\%)} & 0.926 {\color{teal}(+2.6\%)} & 0.936 {\color{teal}(+1.3\%)} \\
\midrule
\multirow{2}{*}{professional\_psychology} & Base & 0.578 & 0.660 & 0.753 & 0.778 & 0.838 & 0.863 \\
 & Role-playing & 0.600 {\color{teal}(+2.1\%)} & 0.655 {\color{red}(-0.5\%)} & 0.752 {\color{red}(-0.2\%)} & 0.784 {\color{teal}(+0.7\%)} & 0.840 {\color{teal}(+0.2\%)} & 0.876 {\color{teal}(+1.3\%)} \\
\midrule
\multirow{2}{*}{public\_relations} & Base & 0.655 & 0.627 & 0.673 & 0.691 & 0.709 & 0.744 \\
 & Role-playing & 0.591 {\color{red}(-6.4\%)} & 0.618 {\color{red}(-0.9\%)} & 0.673 {\color{teal}(+0.0\%)} & 0.745 {\color{teal}(+5.5\%)} & 0.736 {\color{teal}(+2.7\%)} & 0.733 {\color{red}(-1.1\%)} \\
\midrule
\multirow{2}{*}{security\_studies} & Base & 0.653 & 0.694 & 0.694 & 0.735 & 0.771 & 0.803 \\
 & Role-playing & 0.665 {\color{teal}(+1.2\%)} & 0.714 {\color{teal}(+2.0\%)} & 0.698 {\color{teal}(+0.4\%)} & 0.755 {\color{teal}(+2.0\%)} & 0.776 {\color{teal}(+0.4\%)} & 0.816 {\color{teal}(+1.3\%)} \\
\midrule
\multirow{2}{*}{sociology} & Base & 0.761 & 0.766 & 0.806 & 0.826 & 0.841 & 0.863 \\
 & Role-playing & 0.766 {\color{teal}(+0.5\%)} & 0.846 {\color{teal}(+8.0\%)} & 0.826 {\color{teal}(+2.0\%)} & 0.816 {\color{red}(-1.0\%)} & 0.856 {\color{teal}(+1.5\%)} & 0.876 {\color{teal}(+1.3\%)} \\
\midrule
\multirow{2}{*}{us\_foreign\_policy} & Base & 0.790 & 0.860 & 0.810 & 0.830 & 0.860 & 0.883 \\
 & Role-playing & 0.770 {\color{red}(-2.0\%)} & 0.850 {\color{red}(-1.0\%)} & 0.810 {\color{teal}(+0.0\%)} & 0.830 {\color{teal}(+0.0\%)} & 0.880 {\color{teal}(+2.0\%)} & 0.896 {\color{teal}(+1.3\%)} \\
\midrule
\multirow{2}{*}{virology} & Base & 0.512 & 0.512 & 0.512 & 0.566 & 0.554 & 0.584 \\
 & Role-playing & 0.494 {\color{red}(-1.8\%)} & 0.464 {\color{red}(-4.8\%)} & 0.500 {\color{red}(-1.2\%)} & 0.542 {\color{red}(-2.4\%)} & 0.566 {\color{teal}(+1.2\%)} & 0.573 {\color{red}(-1.1\%)} \\
\midrule
\multirow{2}{*}{world\_religions} & Base & 0.754 & 0.784 & 0.819 & 0.854 & 0.883 & 0.893 \\
 & Role-playing & 0.749 {\color{red}(-0.6\%)} & 0.795 {\color{teal}(+1.2\%)} & 0.848 {\color{teal}(+2.9\%)} & 0.871 {\color{teal}(+1.8\%)} & 0.883 {\color{teal}(+0.0\%)} & 0.905 {\color{teal}(+1.2\%)} \\
\bottomrule
\end{tabular}
}
\caption{Performance comparison of base vs. role-playing modes across different models and subjects on MMLU (Part 2/3).}
\label{table:13}
\end{table*}

\renewcommand{\arraystretch}{0.9}
\setlength{\tabcolsep}{3pt}
\begin{table*}[t]
\centering
\footnotesize
\resizebox{\linewidth}{!}{
\begin{tabular}{lccccccc}
\toprule
\multirow{2}{*}{Subject} & \multirow{2}{*}{Setting} & \multicolumn{6}{c}{Models} \\
\cmidrule(lr){3-8}
 & & Mistral-7B & Llama3-8B & Qwen3-4B & Qwen3-8B & Qwen3-14B & Qwen3-32B \\
\midrule
\multirow{2}{*}{nutrition} & Base & 0.650 & 0.725 & 0.778 & 0.840 & 0.876 & 0.893 \\
 & Role-playing & 0.621 {\color{red}(-2.9\%)} & 0.755 {\color{teal}(+2.9\%)} & 0.784 {\color{teal}(+0.7\%)} & 0.827 {\color{red}(-1.3\%)} & 0.866 {\color{red}(-1.0\%)} & 0.905 {\color{teal}(+1.2\%)} \\
\midrule
\multirow{2}{*}{philosophy} & Base & 0.637 & 0.617 & 0.701 & 0.714 & 0.775 & 0.803 \\
 & Role-playing & 0.666 {\color{teal}(+2.9\%)} & 0.672 {\color{teal}(+5.5\%)} & 0.707 {\color{teal}(+0.6\%)} & 0.736 {\color{teal}(+2.3\%)} & 0.752 {\color{red}(-2.3\%)} & 0.814 {\color{teal}(+1.1\%)} \\
\midrule
\multirow{2}{*}{prehistory} & Base & 0.685 & 0.701 & 0.772 & 0.843 & 0.861 & 0.883 \\
 & Role-playing & 0.670 {\color{red}(-1.5\%)} & 0.719 {\color{teal}(+1.9\%)} & 0.762 {\color{red}(-0.9\%)} & 0.830 {\color{red}(-1.2\%)} & 0.877 {\color{teal}(+1.5\%)} & 0.896 {\color{teal}(+1.3\%)} \\
\midrule
\multirow{2}{*}{professional\_accounting} & Base & 0.447 & 0.511 & 0.709 & 0.780 & 0.844 & 0.863 \\
 & Role-playing & 0.447 {\color{teal}(+0.0\%)} & 0.493 {\color{red}(-1.8\%)} & 0.716 {\color{teal}(+0.7\%)} & 0.812 {\color{teal}(+3.2\%)} & 0.812 {\color{red}(-3.2\%)} & 0.875 {\color{teal}(+1.2\%)} \\
\midrule
\multirow{2}{*}{professional\_law} & Base & 0.423 & 0.453 & 0.465 & 0.492 & 0.581 & 0.624 \\
 & Role-playing & 0.427 {\color{teal}(+0.4\%)} & 0.443 {\color{red}(-1.0\%)} & 0.449 {\color{red}(-1.6\%)} & 0.483 {\color{red}(-0.9\%)} & 0.571 {\color{red}(-1.0\%)} & 0.613 {\color{red}(-1.1\%)} \\
\midrule
\multirow{2}{*}{professional\_medicine} & Base & 0.614 & 0.713 & 0.824 & 0.871 & 0.901 & 0.923 \\
 & Role-playing & 0.632 {\color{teal}(+1.8\%)} & 0.695 {\color{red}(-1.8\%)} & 0.805 {\color{red}(-1.8\%)} & 0.849 {\color{red}(-2.2\%)} & 0.926 {\color{teal}(+2.6\%)} & 0.936 {\color{teal}(+1.3\%)} \\
\midrule
\multirow{2}{*}{professional\_psychology} & Base & 0.578 & 0.660 & 0.753 & 0.778 & 0.838 & 0.863 \\
 & Role-playing & 0.600 {\color{teal}(+2.1\%)} & 0.655 {\color{red}(-0.5\%)} & 0.752 {\color{red}(-0.2\%)} & 0.784 {\color{teal}(+0.7\%)} & 0.840 {\color{teal}(+0.2\%)} & 0.876 {\color{teal}(+1.3\%)} \\
\midrule
\multirow{2}{*}{public\_relations} & Base & 0.655 & 0.627 & 0.673 & 0.691 & 0.709 & 0.744 \\
 & Role-playing & 0.591 {\color{red}(-6.4\%)} & 0.618 {\color{red}(-0.9\%)} & 0.673 {\color{teal}(+0.0\%)} & 0.745 {\color{teal}(+5.5\%)} & 0.736 {\color{teal}(+2.7\%)} & 0.733 {\color{red}(-1.1\%)} \\
\midrule
\multirow{2}{*}{security\_studies} & Base & 0.653 & 0.694 & 0.694 & 0.735 & 0.771 & 0.803 \\
 & Role-playing & 0.665 {\color{teal}(+1.2\%)} & 0.714 {\color{teal}(+2.0\%)} & 0.698 {\color{teal}(+0.4\%)} & 0.755 {\color{teal}(+2.0\%)} & 0.776 {\color{teal}(+0.4\%)} & 0.816 {\color{teal}(+1.3\%)} \\
\midrule
\multirow{2}{*}{sociology} & Base & 0.761 & 0.766 & 0.806 & 0.826 & 0.841 & 0.863 \\
 & Role-playing & 0.766 {\color{teal}(+0.5\%)} & 0.846 {\color{teal}(+8.0\%)} & 0.826 {\color{teal}(+2.0\%)} & 0.816 {\color{red}(-1.0\%)} & 0.856 {\color{teal}(+1.5\%)} & 0.876 {\color{teal}(+1.3\%)} \\
\midrule
\multirow{2}{*}{us\_foreign\_policy} & Base & 0.790 & 0.860 & 0.810 & 0.830 & 0.860 & 0.883 \\
 & Role-playing & 0.770 {\color{red}(-2.0\%)} & 0.850 {\color{red}(-1.0\%)} & 0.810 {\color{teal}(+0.0\%)} & 0.830 {\color{teal}(+0.0\%)} & 0.880 {\color{teal}(+2.0\%)} & 0.896 {\color{teal}(+1.3\%)} \\
\midrule
\multirow{2}{*}{virology} & Base & 0.512 & 0.512 & 0.512 & 0.566 & 0.554 & 0.584 \\
 & Role-playing & 0.494 {\color{red}(-1.8\%)} & 0.464 {\color{red}(-4.8\%)} & 0.500 {\color{red}(-1.2\%)} & 0.542 {\color{red}(-2.4\%)} & 0.566 {\color{teal}(+1.2\%)} & 0.573 {\color{red}(-1.1\%)} \\
\midrule
\multirow{2}{*}{world\_religions} & Base & 0.754 & 0.784 & 0.819 & 0.854 & 0.883 & 0.893 \\
 & Role-playing & 0.749 {\color{red}(-0.6\%)} & 0.795 {\color{teal}(+1.2\%)} & 0.848 {\color{teal}(+2.9\%)} & 0.871 {\color{teal}(+1.8\%)} & 0.883 {\color{teal}(+0.0\%)} & 0.905 {\color{teal}(+1.2\%)} \\
\bottomrule
\end{tabular}
}
\caption{Performance comparison of base vs. role-playing modes across different models and subjects on MMLU (Part 3/3).}
\label{table:14}
\end{table*}

\begin{table*}[t]
\centering
\caption{Performance of SLVP and SLCP on MMLU-Redux and MMLU benchmarks across 8 languages. For each row, the \colorbox{lightblue}{best} result is highlighted.}

\label{table:15}
\resizebox{\linewidth}{!}{
\begin{tabular}{lllcccccccc|c}
\toprule
Model & Benchmark & Variant & en & zh & ja & km & ms & de & es & fr & avg \\
\midrule
\multirow{4}{*}{Qwen3-4B}
& \multirow{2}{*}{MMLU-Redux} & SLVP & 0.741 & 0.730 & 0.751 & 0.729 & 0.751 & 0.758 & \colorbox{lightblue}{0.764} & 0.760 & 0.748 \\
&& SLCP & 0.774 & 0.752 & 0.773 & 0.762 & 0.775 & \colorbox{lightblue}{0.782} & 0.778 & \colorbox{lightblue}{0.782} & 0.772 \\
\cmidrule(lr){2-12}
& \multirow{2}{*}{MMLU} & SLVP & 0.760 & 0.738 & 0.769 & 0.765 & 0.770 & 0.772 & \colorbox{lightblue}{0.773} & \colorbox{lightblue}{0.773} & 0.765 \\
&& SLCP & 0.776 & 0.751 & \colorbox{lightblue}{0.779} & 0.771 & 0.776 & \colorbox{lightblue}{0.779} & 0.776 & 0.778 & 0.773 \\
\midrule
\multirow{4}{*}{Qwen3-8B}
& \multirow{2}{*}{MMLU-Redux} & SLVP & 0.755 & 0.751 & 0.769 & 0.754 & 0.773 & 0.774 & \colorbox{lightblue}{0.779} & 0.776 & 0.766 \\
&& SLCP & \colorbox{lightblue}{0.819} & 0.781 & 0.811 & 0.805 & 0.814 & 0.815 & \colorbox{lightblue}{0.819} & \colorbox{lightblue}{0.819} & 0.810 \\
\cmidrule(lr){2-12}
& \multirow{2}{*}{MMLU} & SLVP & 0.788 & 0.779 & 0.794 & 0.794 & 0.796 & 0.798 & \colorbox{lightblue}{0.801} & 0.797 & 0.793 \\
&& SLCP & 0.812 & 0.790 & 0.809 & 0.809 & \colorbox{lightblue}{0.815} & 0.812 & 0.811 & 0.813 & 0.809 \\
\midrule
\multirow{4}{*}{Qwen3-14B}
& \multirow{2}{*}{MMLU-Redux} & SLVP & 0.803 & 0.797 & 0.816 & 0.793 & 0.810 & 0.817 & \colorbox{lightblue}{0.825} & 0.815 & 0.810 \\
&& SLCP & \colorbox{lightblue}{0.841} & 0.829 & 0.840 & 0.837 & 0.838 & 0.839 & 0.839 & \colorbox{lightblue}{0.841} & 0.838 \\
\cmidrule(lr){2-12}
& \multirow{2}{*}{MMLU} & SLVP & 0.825 & 0.813 & 0.831 & 0.826 & 0.830 & 0.832 & 0.834 & \colorbox{lightblue}{0.836} & 0.828 \\
&& SLCP & 0.840 & 0.829 & 0.841 & 0.840 & 0.841 & \colorbox{lightblue}{0.844} & 0.841 & 0.840 & 0.840 \\
\midrule
\multirow{4}{*}{Qwen3-32B}
& \multirow{2}{*}{MMLU-Redux} & SLVP & 0.821 & 0.814 & 0.833 & 0.811 & 0.828 & 0.835 & \colorbox{lightblue}{0.849} & 0.832 & 0.828 \\
&& SLCP & \colorbox{lightblue}{0.865} & 0.846 & 0.856 & 0.853 & 0.855 & 0.856 & 0.856 & 0.857 & 0.856 \\
\cmidrule(lr){2-12}
& \multirow{2}{*}{MMLU} & SLVP & 0.842 & 0.830 & 0.848 & 0.843 & 0.847 & 0.849 & 0.851 & \colorbox{lightblue}{0.856} & 0.846 \\
&& SLCP & 0.856 & 0.846 & 0.857 & 0.856 & 0.857 & \colorbox{lightblue}{0.862} & 0.857 & 0.856 & 0.856 \\
\midrule
\multirow{4}{*}{Llama3-8B}
& \multirow{2}{*}{MMLU-Redux} & SLVP & 0.523 & 0.523 & 0.514 & 0.309 & 0.527 & 0.528 & 0.487 & \colorbox{lightblue}{0.529} & 0.493 \\
&& SLCP & 0.470 & 0.510 & 0.487 & 0.405 & 0.498 & \colorbox{lightblue}{0.522} & 0.427 & 0.466 & 0.473 \\
\cmidrule(lr){2-12}
& \multirow{2}{*}{MMLU} & SLVP & 0.598 & 0.602 & 0.586 & 0.324 & 0.612 & 0.607 & 0.545 & \colorbox{lightblue}{0.614} & 0.561 \\
&& SLCP & 0.525 & 0.583 & 0.544 & 0.446 & 0.563 & \colorbox{lightblue}{0.594} & 0.473 & 0.512 & 0.530 \\
\midrule
\multirow{4}{*}{Mistral-7B}
& \multirow{2}{*}{MMLU-Redux} & SLVP & 0.446 & \colorbox{lightblue}{0.473} & 0.472 & 0.416 & 0.454 & \colorbox{lightblue}{0.473} & 0.446 & \colorbox{lightblue}{0.473} & 0.457 \\
&& SLCP & 0.416 & 0.465 & 0.398 & 0.379 & \colorbox{lightblue}{0.471} & 0.465 & 0.427 & 0.415 & 0.430 \\
\cmidrule(lr){2-12}
& \multirow{2}{*}{MMLU} & SLVP & 0.548 & 0.584 & 0.575 & 0.505 & 0.558 & \colorbox{lightblue}{0.591} & 0.552 & 0.585 & 0.562 \\
&& SLCP & 0.458 & 0.511 & 0.392 & 0.410 & 0.517 & \colorbox{lightblue}{0.576} & 0.520 & 0.552 & 0.492 \\
\bottomrule
\end{tabular}
}
\end{table*}

\end{document}